\documentclass[onecolumn,10pt,a4paper]{article}
\usepackage{amsfonts}
\usepackage{algorithm}
\usepackage{algorithmic}
\usepackage[ansinew]{inputenc}

\usepackage[pdftex]{graphicx}
\usepackage[cmex10]{amsmath}
\usepackage{url}

\newtheorem{thm}{Theorem}

\newtheorem{ass}{Assumption}
\newtheorem{lem}{Lemma}

\def\urltilda{\kern -.15em\lower .7ex\hbox{\~{}}\kern .04em}
\def\urldot{\kern -.10em.\kern -.10em}
\def\urlhttp{http\kern -.10em\lower -.1ex\hbox{:}\kern -.12em\lower 0ex\hbox{/}\kern -.18em\lower 0ex\hbox{/}}

\begin{document}
%
\title{Client-server multi-task learning from distributed datasets}
%
%
%
\date{}
\author{Francesco~Dinuzzo\thanks{Francesco~Dinuzzo is with Department of Mathematics, University of Pavia, and  Risk and Security Study Center, Istituto Universitario di Studi Superiori (IUSS), Pavia, Italy e-mail: francesco.dinuzzo@unipv.it.} ,
        Gianluigi~Pillonetto\thanks{Gianluigi~Pillonetto is with Department of Information Engineering, University of Padova, Padova, Italy, e-mail: giapi@dei.unipd.it} ,
        and Giuseppe~De~Nicolao\thanks{Giuseppe~De~Nicolao is with Department of Computer Engineering and Systems Science, University of Pavia, Pavia, Italy, e-mail: giuseppe.denicolao@unipv.it}
}
\maketitle

\begin{abstract}
A client-server architecture to simultaneously solve multiple learning tasks from distributed datasets is described. In such architecture, each client is associated with an individual learning task and the associated dataset of examples. The goal of the architecture is to perform information fusion from multiple datasets while preserving privacy of individual data. The role of the server is to collect data in real-time from the clients and codify the information in a common database. The information coded in this database can be used by all the clients to solve their individual learning task, so that each client can exploit the informative content of all the datasets without actually having access to private data of others. The proposed algorithmic framework, based on regularization theory and kernel methods, uses a suitable class of “mixed effect” kernels. The new method is illustrated through a simulated music recommendation system.
\end{abstract}

\section{Introduction} \label{sec01}
%
%
%
%

The solution of learning tasks by joint analysis of multiple datasets is receiving increasing attention in different fields and under various perspectives. Indeed, the information provided by data for a specific task may serve as a domain-specific inductive bias for the others. Combining datasets to solve multiple learning tasks is an approach known in the machine learning literature as \emph{multi-task learning} or \emph{learning to learn} \cite{Thrun96, Caruana97, Thrun97, Baxter97, Ben-david02, Bakker03,  Lawrence04}. In this context, the analysis of the \emph{inductive transfer} process and the investigation of general methodologies for the simultaneous learning of multiple tasks are important topics of research. Many theoretical and experimental results support the intuition that, when relationships exist between the tasks, simultaneous learning performs better than separate (single-task) learning \cite{Schwaighofer05, Yu05, Yu07,  Xue07, Bonilla07, Argyriou07, Bickel08,  Zhang08, Qi08, An08}. Theoretical results include the extension to the multi-task setting of generalization bounds and the notion of VC-dimension \cite{Baxter00, Ben-david03, Maurer06} and a methodology for learning multiple tasks exploiting unlabeled data (the so-called semi-supervised setting) \cite{Ando05}.

Importance of combining datasets is especially evident in biomedicine. In pharmacological experiments, few training examples are typically available for a specific subject due to technological and ethical constraints \cite{Carson83,Jacquez85}. This makes hard to formulate and quantify models from experimental data. To obviate this problem, the so-called \emph{population method} has been studied and applied with success since the seventies in pharmacology \cite{Sheiner77, Beal82, Yuh94}. Population methods are based on the knowledge that subjects, albeit different, belong to a population of similar individuals, so that data collected in one subject may be informative with respect to the others \cite{Vozeh96,Park97}. Such population approaches belongs to the family of so-called \emph{mixed-effect} statistical methods. In these methods, clinical measurements from different subjects are combined to simultaneously learn individual features of physiological responses to drug administration \cite{Sheiner00}. Population methods have been applied with success also in other biomedical contexts such as medical imaging and bioinformatics \cite{Ferrazzi03,Bertoldo04}. Classical approaches postulate finite-dimensional nonlinear dynamical systems whose unknown parameters can be determined by means of optimization algorithms \cite{Beal92,Sheiner94,Davidian95,Aarons99}. Other strategies include Bayesian estimation with stochastic simulation \cite{Wakefield94,Lunn02,Gilks96} and nonparametric population methods \cite{Fattinger95,Magni02,Neve05b, Neve07, Neve08, Pillonetto3}.

Information fusion from different but related datasets is widespread also in econometrics and marketing analysis, where the goal is to learn user preferences by analyzing both user-specific information and information from related users, see e.g. \cite{Srivastava71, Arora98, Allenby99, Greene02}. The so-called \emph{conjoint analysis} aims to determine the features of a product that mostly influence customer's decisions. In the web, collaborative approaches to estimate user preferences have become standard methodologies in many commercial systems and social networks, under the name of \emph{collaborative filtering} or \emph{recommender systems}, see e.g. \cite{Resnick97}. Pioneering collaborative filtering systems include Tapestry \cite{Goldberg92}, GroupLens \cite{Resnick94, Konstan97}, ReferralWeb \cite{Kautz97}, PHOAKS \cite{Terveen97}. More recently, the collaborative filtering problem has been attacked with machine learning methodologies such as Bayesian networks \cite{Breese98}, MCMC algorithms \cite{Chen99}, mixture models \cite{Hofmann99}, dependency networks \cite{Heckerman00}, maximum margin matrix factorization \cite{Srebro05}.

Coming back to the machine learning literature, in the single-task context much attention has been given in the last years to non-parametric techniques such as kernel methods \cite{Scholkopf01b} and Gaussian processes \cite{Rasmussen06}. These approaches are powerful and theoretically sound, having their mathematical foundations in regularization theory for inverse problems, statistical learning theory and Bayesian estimation \cite{Aronszajn50, Tikhonov77, Poggio90, Wahba90, Vapnik98, Cucker01}. The flexibility of kernel engineering allows for the estimation of functions defined on generic sets from arbitrary sources of data. These methodologies have been recently extended to the multi-task setting. In \cite{Evgeniou05}, a general framework to solve multi-task learning problems using kernel methods and regularization has been proposed, relying on the theory of reproducing kernel Hilbert spaces (RKHS) of vector-valued functions \cite{Micchelli05}.

In many applications (e-commerce, social network data processing, recommender systems), real-time processing of examples is required. On-line multi-task learning schemes find their natural application in data mining problems involving very large datasets, and are therefore required to scale well with the number of tasks and examples. In \cite{Pillonetto4}, an on-line task-wise algorithm to solve multi-task regression problems has been proposed. The learning problem is formulated in the context of on-line Bayesian estimation, see e.g. \cite{Opper98, Csato02}, within which Gaussian processes with suitable covariance functions are used to characterize a non-parametric mixed-effect model. One of the key features of the algorithm is the capability to exploit shared inputs between the tasks in order to reduce computational complexity. However, the algorithm in \cite{Pillonetto4} has a centralized structure in which tasks are sequentially analyzed, and is not able to address neither architectural issues regarding the flux of information nor privacy protection.

In this paper, multi-task learning from distributed datasets is addressed using a client-server architecture. In our scheme, clients are in a one-to-one correspondence with tasks and their individual database of examples. The role of the server is to collect examples from different clients in order to summarize their informative content. When a new example associated with any task becomes available, the server executes an on-line update algorithm. While in \cite{Pillonetto4} different tasks are sequentially analyzed, the architecture presented in this paper can process examples coming in any order from different learning tasks. The summarized information is stored in a \emph{disclosed database} whose content is available for download enabling each client to compute its own estimate exploiting the informative content of all the other datasets. Particular attention is paid to confidentiality issues, especially valuable in commercial and recommender systems, see e.g. \cite{Ramakrishnan01, Canny02}. First, we require that each specific client cannot access other clients data. In addition, individual datasets cannot be reconstructed from the disclosed database. Two kind of clients are considered: \emph{active} and \emph{passive} ones. An active client sends its data to the server, thus contributing to the collaborative estimate. A passive client only downloads information from the disclosed database without sending its data. A regularization problem with a parametric bias term is considered in which a \emph{mixed-effect kernel} is used to exploit relationships between the tasks. Albeit specific, the mixed-effect non-parametric model is quite flexible, and its usefulness has been demonstrated in several works \cite{Neve05b, Neve07, Lu08, Pillonetto4}.

The paper is organized as follows. Multi-task learning with regularized kernel methods is presented in section \ref{sec02}, in which a class of mixed-effect kernels is also introduced. In section \ref{sec03}, an efficient centralized off-line algorithm for multi-task learning is described that solves the regularization problem of section \ref{sec02}. In section \ref{sec04}, a rather general client-server architecture is described, which is able to efficiently solve online multi-task learning from distributed datasets. The server-side algorithm is derived and discussed in subsection \ref{sub04-1}, while the client-side algorithm for both active and passive clients is derived in subsection \ref{sub04-2}. In section \ref{sec05}, a simulated music recommendation system is employed to test performances of our algorithm. Conclusions (section \ref{sec06}) end the paper. The Appendix contains technical lemmas and proofs.

\subsection*{Notational preliminaries}

\begin{itemize}
  \item $X$ denotes a generic set with cardinality $|X|$.
  \item A \emph{vector} is an element of $\mathbf{a} \in X^{n}$ (an object with one index). Vectors are denoted by lowercase bold characters. Vector components are denoted by the corresponding non-bold letter with a subscript (e.g. $a_i$ denotes the $i$-th component of $\mathbf{a}$).
  \item A \emph{matrix} is an element of $\mathbf{A} \in X^{n \times m}$ (an object with two indices). Matrices are denoted by uppercase bold characters.      Matrix entries are denoted by the corresponding non-bold letter with two subscript (e.g. $A_{ij}$ denotes the entry of place $(i,j)$ of $\mathbf{A}$).
  \item Vectors $\mathbf{y} \in \mathbb{R}^{n}$ are associated with column matrices, unless otherwise specified.
  \item For all $n \in \mathbb{N}$, let $[n] := \{1, 2, \ldots, n\}$.
  \item Let $\mathbf{I}$ denote the identity matrix of suitable dimension.
  \item Let $\mathbf{e}_i \in \mathbb{R}^n$ denote the $i$-th element of the canonical basis of $\mathbb{R}^n$ (all zeros with 1 in position $i$):
    \[
    \mathbf{e}_i := \left(%
    \begin{array}{ccccc}
    0 & \cdots & 1 & \cdots & 0 \\
    \end{array}%
    \right)^T.
    \]
   \item An $(n,p)$ \emph{index vector} is an object $\mathbf{k} \in [n]^{p}$.
   \item Given a vector $\mathbf{a} \in X^{n}$ and an $(n,p)$ index vector $\mathbf{k}$, let
   \[
    \mathbf{a}(\mathbf{k}) := \left(
                           \begin{array}{ccc}
                            a_{k_1} & \cdots & a_{k_p} \\
                           \end{array}
                           \right) \in X^{p}.
    \]
   \item Given a matrix $\mathbf{A} \in X^{n \times m}$ and two index vectors $\mathbf{k}^1$ e $\mathbf{k}^2$, that are $(n,p_1)$ and $(m,p_2)$, respectively, let
    \[
    \mathbf{A}(\mathbf{k}^1,\mathbf{k}^2) := \left(
                                          \begin{array}{ccc}
                                           A_{k^1_1 k^2_1}  & \cdots & A_{k^1_1 k^2_{p_2}} \\
                                            \vdots & \ddots & \vdots \\
                                           A_{k^1_{p_1} k^2_1}  & \cdots & A_{k^1_{p_1} k^2_{p_2}} \\
                                          \end{array}
                                        \right) \in X^{p_1 \times p_2}.
    \]
    \item Finally, let
    \[
    \mathbf{A}(:,\mathbf{k}^2) := \mathbf{A}([n],\mathbf{k}^2),\quad \mathbf{A}(\mathbf{k}^1,:):=\mathbf{A}(\mathbf{k}^1, [m]).
    \]
\end{itemize}

Notice that vectors, as defined in this paper, are not necessarily elements of a vector space. The definition of “vector” adopted in this paper is similar to that used in standard object-oriented programming languages such as C++.

\section{Problem formulation} \label{sec02}

Let $m \in \mathbb{N}$ denote the total number of tasks. For the task $j$, a vector of $\ell_j$ input-output pairs $\mathbf{S}^j \in\left(X\times \mathbb{R}\right)^{\ell_j}$ is available:
\[
\mathbf{S}^j := \left(
                  \begin{array}{ccc}
                    (x_{1j}, y_{1j}) & \cdots & (x_{\ell_j j}, y_{\ell_j j}) \\
                  \end{array}
                \right),
\]

\noindent sampled from a distribution $P_j$ on $X \times\mathbb{R}$. The aim of a multi-task regression problem is learning $m$ functions $f_j : X \rightarrow \mathbb{R}$, such that expected errors with respect to some \emph{loss function} $L$
\[
\int_{X \times \mathbb{R}}L(y,f_j(x)) dP_j
\]
\noindent are small.

\emph{Task-labeling} is a simple technique to reduce multi-task learning problems to single-task ones. Task-labels are integers $t_i$ that identify a specific task, say $t_i \in [m]$. The overall dataset can be viewed as a set of triples $\mathbf{S} \in \left([m] \times X \times \mathbb{R}\right)^{\ell}$, where $\ell := \sum_{j=1}^{m}\ell_j$ is the overall number of examples:
\[
\mathbf{S} := \left(
                \begin{array}{ccc}
                  (t_{1}, x_{1}, y_{1}) & \cdots & (t_{\ell}, x_{\ell}, y_{\ell}) \\
                \end{array}
              \right).
\]

\noindent Thus, we can learn a single scalar-valued function defined over an input space enlarged with the task-labels $f: X \times [m] \rightarrow \mathbb{R}$. The correspondence between the dataset $\mathbf{S}_j$ and $\mathbf{S}$ is recovered through an $(\ell,\ell_j)$ index vector $\mathbf{k}^j$  such that
\[
t_{k^j_i} = j, \quad i \in [\ell_j].
\]

Let $\mathcal{H}$ denote an RKHS of functions defined on the enlarged input space $X \times [m]$ with kernel $K$, and $\mathcal{B}$ denote a $d$-dimensional \emph{bias}-space. Solving the multi-task learning problem by regularized kernel methods amounts to finding $\hat{f} \in \mathcal{H}+\mathcal{B}$, such that
\begin{equation} \label{E01}
\hat{f} = \arg \min_{f \in \mathcal{H}+\mathcal{B}} \left( \sum_{i=1}^{\ell}V_i(y_i, f(x_i,t_i) ) + \frac{\lambda}{2}\|P_{\mathcal{H}} f \|^2 _{\mathcal{H}} \right),
\end{equation}

\noindent where $V_i:\mathbb{R}\times\mathbb{R} \rightarrow \mathbb{R}$ are suitable loss functions, $\lambda \geq 0$ is the regularization parameter and $P_{\mathcal{H}}$ is the projection operator into $\mathcal{H}$. In this paper, the focus is on the \emph{mixed effect kernels}, with the following structure:
\begin{multline}\label{EQ100}
K(x_1,t_1,x_2,t_2) = \alpha \overline{K}(x_1,x_2) \\ + (1-\alpha) \sum_{j=1}^{m}  K_T^{j}(t_1,t_2) \widetilde{K}^{j}(x_1,x_2).
\end{multline}
\noindent where
\[
0 \leq \alpha \leq 1.
\]
\noindent Kernels $\overline{K}$ and $\widetilde{K}^{j}$ are defined on $X \times X$ and can possibly be all distinct. On the other hand, $K_T^{j}$ are “selector” kernels defined on $[m] \times [m]$ as
\[
K_T^{j}(t_1,t_2) = \left\{%
\begin{array}{ll}
    1, & t_1 = t_2 = j; \\
    0, & \hbox{otherwhise} \\
\end{array}%
\right.
\]

\noindent Kernels $K_T^{j}$ are not strictly positive. Assume that $\mathcal{B}$ is spanned by functions $\{\alpha \psi_i\}_{1}^{d}$. Of course, this is the same of using $\{\psi_i\}_{1}^{d}$ when $\alpha \neq 0$. However, weighting the basis functions by $\alpha$ is convenient to recover the separate approach ($\alpha = 0$) by continuity. Usually, the dimension $d$ is relatively low. A common choice might be $d=1$ with $\psi_1 = \alpha$, that is $\mathcal{B}$ is simply the space of constant functions, useful to make the learning method translation-invariant.

Under rather general hypotheses for the loss function $V$, the representer theorem, see e.g. \cite{Kimeldorf71}, \cite{Scholkopf01} gives the following expression for the optimum $\hat{f}$:
\begin{align*}
\hat{f}(x,t) = & \alpha \left(\sum_{i=1}^{\ell}a_i \overline{K}(x_i,x) +\sum_{i=1}^{d}b_i\psi_{i}(x)\right)\\
 &+ (1-\alpha) \sum_{i=1}^{\ell}\sum_{j=1}^{m}a_i K_{T}^{j}(t_i,t) \widetilde{K}^{j}(x_i,x).
\end{align*}

\noindent The estimate $\hat{f}_j$ is defined to be the function obtained by plugging the corresponding task-label $t = j$ in the previous expression. As a consequence of the structure of $K$, the expression of $\hat{f}_j$ decouples into two parts:
\begin{equation}\label{E02}
\hat{f}_j(x) := \hat{f}(x,j) =  \bar{f}(x) + \tilde{f}_j(x),
\end{equation}
\noindent where
\begin{align*}
\bar{f}(x) & = \alpha \left(\sum_{i=1}^{\ell}a_i \overline{K}(x_i,x)+\sum_{i=1}^{d}b_i\psi_{i}(x)\right),\\
\tilde{f}_j(x) & = (1-\alpha)\sum_{i \in \mathbf{k}^j}^{\ell}a_i \widetilde{K}^{j}(x_i,x).
\end{align*}

\noindent Function $\bar{f}$ is independent of $j$ and can be regarded as a sort of \emph{average task}, whereas $\tilde{f}_j$ is a non-parametric \emph{individual shift}. The value $\alpha$ is related to the “shrinking” of the individual estimates toward the average task. When $\alpha = 1$, the same function is learned for all the tasks, as if all examples referred to an unique task (\emph{pooled} approach). On the other hand, when $\alpha= 0$, all the tasks are learned independently (\emph{separate} approach), as if tasks were not related at all.

Throughout this paper, the problem is specialized to the case of (weighted) squared loss functions
\[
V_i(y,f(x,t)) = \frac{1}{2w_i}(y-f(x,t))^2,
\]

\noindent where $\mathbf{w} \in \mathbb{R}_+^{\ell}$ denote a weight vector. For squared loss functions, coefficient vectors $\mathbf{a}$ and $\mathbf{b}$ can be obtained by solving the linear system \cite{Wahba90}
\begin{equation}\label{E03}
\left(%
\begin{array}{cc}
  \mathbf{K} +\lambda\mathbf{W}& \boldsymbol{\Psi} \\
  \boldsymbol{\Psi}^T & \mathbf{0} \\
\end{array}%
\right)\left(%
\begin{array}{c}
  \mathbf{a} \\
  \alpha \mathbf{b} \\
\end{array}%
\right) = \left(%
\begin{array}{c}
  \mathbf{y} \\
  \mathbf{0} \\
\end{array}%
\right)
\end{equation}

\noindent where
\[
\mathbf{W} = \textrm{diag}(\mathbf{w}),
\]
\[
\mathbf{K} =\left(\alpha \mathbf{\overline{K}}+(1-\alpha)\sum_{j=1}^{m}\mathbf{I}(:,\mathbf{k}^j) \mathbf{\widetilde{K}}^j(\mathbf{k}^j,\mathbf{k}^j)\mathbf{I}(\mathbf{k}^j,:)\right),
\]
\[
\overline{K}_{ij} = \overline{K}(x_i,x_j), \quad \widetilde{K}^k_{ij} = \widetilde{K}^k(x_i,x_j), \quad \Psi_{ij} =\psi_{j}(x_i).
\]

\noindent For $\alpha =0$, vector $\mathbf{b}$ is not well determined. The linear system can be also solved via \emph{back-fitting} on the residual generated by the parametric bias estimate:
\begin{eqnarray}
\label{E04}\alpha\left[\boldsymbol{\Psi}^T\left(\mathbf{K}+\lambda \mathbf{W}\right)^{-1}\boldsymbol{\Psi}\right]\mathbf{b} & =& \boldsymbol{\Psi}^T\left(\mathbf{K}+\lambda \mathbf{W}\right)^{-1}\mathbf{y},\\
\label{E05}\left(\mathbf{K} +\lambda \mathbf{W}\right)\mathbf{a} & =& \mathbf{y}-\alpha\boldsymbol{\Psi}\mathbf{b}.
\end{eqnarray}

\section{Complexity reduction} \label{sec03}

In many applications of multi-task learning, some or all of the input data $x_i$ are shared between the tasks so that the number of different basis functions appearing in the expansion (\ref{E02}) may be considerably less than $\ell$. As explained below, this feature can be exploited to derive efficient incremental online algorithms for multi-task learning. Introduce the vector of \emph{unique inputs}
\[
\breve{\mathbf{x}} \in X^{n}, \quad \hbox{ such that } \quad \breve{x}_i \neq \breve{x}_j, \quad \forall i \neq j,
\]
\noindent where $n < \ell$ denote the number of unique inputs. For each task $j$, a new $(n,\ell_j)$ index vector $\mathbf{h}^j$ can be defined such that
\[
x_{ij} = \breve{x}_{h^j_i} , \quad i \in [\ell_j].
\]

\noindent Let also
\[
\mathbf{a}^j := \mathbf{a}(\mathbf{k}^j), \quad \mathbf{y}^{j} := \mathbf{y}(\mathbf{k}^j), \quad \mathbf{w}^j := \mathbf{w}(\mathbf{k}^j).
\]

\noindent The information contained in the index vectors $\mathbf{h}^j$ is equivalently represented through a binary matrix  $\mathbf{P} \in \{0, 1\}^{\ell \times n}$, such that
\[\mathbf{P}_{ij} = \left\{
                    \begin{array}{ll}
                      1, & x_i = \breve{x}_j \\
                      0, & \hbox{otherwise.}
                    \end{array}
                  \right.
\]

\noindent We have the following decompositions:
\begin{equation}\label{E06}
\breve{\mathbf{a}} := \mathbf{P}^T\mathbf{a}, \quad \mathbf{\overline{K}} = \mathbf{P}\breve{\mathbf{K}} \mathbf{P}^T, \quad \breve{\mathbf{K}} := \mathbf{L}\mathbf{D}\mathbf{L}^T, \quad \boldsymbol{\Psi} = \mathbf{P}\breve{\boldsymbol{\Psi}},
\end{equation}

\noindent where $\mathbf{L} \in \mathbb{R}^{n\times r}$, $\mathbf{D} \in \mathbb{R}^{r\times r}$ are suitable rank-$r$ factors, $\mathbf{D}$ is diagonal, and $\mathbf{L} \mathbf{D}\mathbf{L}^T \in \mathbb{R}^{n \times n}$. $\breve{\mathbf{K}} \in \mathbb{R}^{n \times n}$ is a kernel matrix associated with the condensed input set $\breve{\mathbf{x}}$: $\breve{K}_{ij} = \overline{K}(\breve{x}_i,\breve{x}_j)$, $\breve{\boldsymbol{\Psi}} \in \mathbb{R}^{n \times d}$, $\breve{\Psi}_{ij} = \psi_j(\breve{x}_i)$. If  $\overline{K}$ is strictly positive, we can assume $r = n$ and $\mathbf{L}$ can be taken as a lower triangular matrix, see e.g. \cite{Golub96}.

Solution (\ref{E02}) can be rewritten in a compact form:
\begin{align*}
\hat{f}_j(x)  = & \alpha \left(\sum_{i=1}^{n}\breve{a}_i \overline{K}(\breve{x}_i,x)+ \sum_{i=1}^{d}b_i\psi_{i}(x)\right)\\
& +  (1-\alpha) \sum_{i=1}^{\ell_j}a_i^j \widetilde{K}^{j}(\breve{x}_{h^j_i},x).
\end{align*}

Introduce the following “compatibility condition” between kernel $\overline{K}$ and the bias space $\mathcal{B}$.

\begin{ass}\label{ASS1}
There exists  $\mathbf{M} \in \mathbb{R}^{r\times d}$ such that
\[
\mathbf{L}\mathbf{D}\mathbf{M} = \breve{\boldsymbol{\Psi}}.
\]
\end{ass}

\noindent Assumption \ref{ASS1} is automatically satisfied in the no-bias case or when $\overline{K}$ is strictly positive.

The next result shows that coefficients $\mathbf{a}$ and $\mathbf{b}$ can be obtained by solving a system of linear equations involving only “small-sized” matrices so that complexity depends on the number of unique inputs rather then the total number of examples.

\begin{thm} \label{THM1} Let Assumption \ref{ASS1} hold. Coefficient vectors $\mathbf{a}$ and $\mathbf{b}$ can be evaluated through Algorithm \ref{ALG1}. For $\alpha = 0$, $\mathbf{b}$ is undetermined.

\begin{algorithm}[h]
\caption{Centralized off-line algorithm.} \label{ALG1}
\begin{algorithmic}[1]
\STATE $\mathbf{R} \leftarrow \mathbf{0}$
\FOR{$j =1:m$}
\STATE $\mathbf{R}^j \leftarrow \left[ (1-\alpha)\mathbf{\widetilde{K}}^j(\mathbf{k}^j,\mathbf{k}^j) + \lambda \mathbf{W}(\mathbf{k}^j,\mathbf{k}^j) \right]^{-1}$
\STATE $\mathbf{R} \leftarrow \mathbf{R} + \mathbf{I}([\ell],\mathbf{k}^j)\mathbf{R}^j\mathbf{I}(\mathbf{k}^j,[\ell])$
\ENDFOR
\IF{$\alpha \neq 0$}
\STATE $\hbox{Compute factors } \mathbf{L}, \mathbf{D}, \mathbf{M}$
\STATE $\breve{\mathbf{y}} \leftarrow \mathbf{L}^T\mathbf{P}^T\mathbf{R}\mathbf{y}$
\STATE $\mathbf{H} \leftarrow \left(\mathbf{D}^{-1}+\alpha \mathbf{L}^T\mathbf{P}^T\mathbf{R}\mathbf{P}\mathbf{L} \right)^{-1}$
\STATE $\mathbf{b} \leftarrow \hbox{Solution to} \left(\mathbf{M}^T(\mathbf{D}- \mathbf{H})\mathbf{M}\right)\mathbf{b}  = \mathbf{M}^T\mathbf{H}\breve{\mathbf{y}}$
\STATE $\mathbf{a} \leftarrow \mathbf{R}\left[\mathbf{y}-\alpha\mathbf{P}\mathbf{L}\mathbf{H}\left(\breve{\mathbf{y}}+\mathbf{M}\mathbf{b}\right)\right]$
\ELSE
\STATE $\mathbf{a} = \mathbf{R}\mathbf{y}$
\ENDIF
\end{algorithmic}
\end{algorithm}
\end{thm}

Algorithm \ref{ALG1} is an off-line (centralized) procedure whose computational complexity scales with $O(n^3 m+d^3)$. In the following section, a client-server on-line version of Algorithm \ref{ALG1} will be derived that preserves this complexity bound. Typically, this is much better than $O\left((\ell+d)^3\right)$, the worst-case complexity of directly solving (\ref{E03}).

\section{A client-server online algorithm} \label{sec04}

\begin{figure}
\centering
\includegraphics[width=1.0\linewidth]{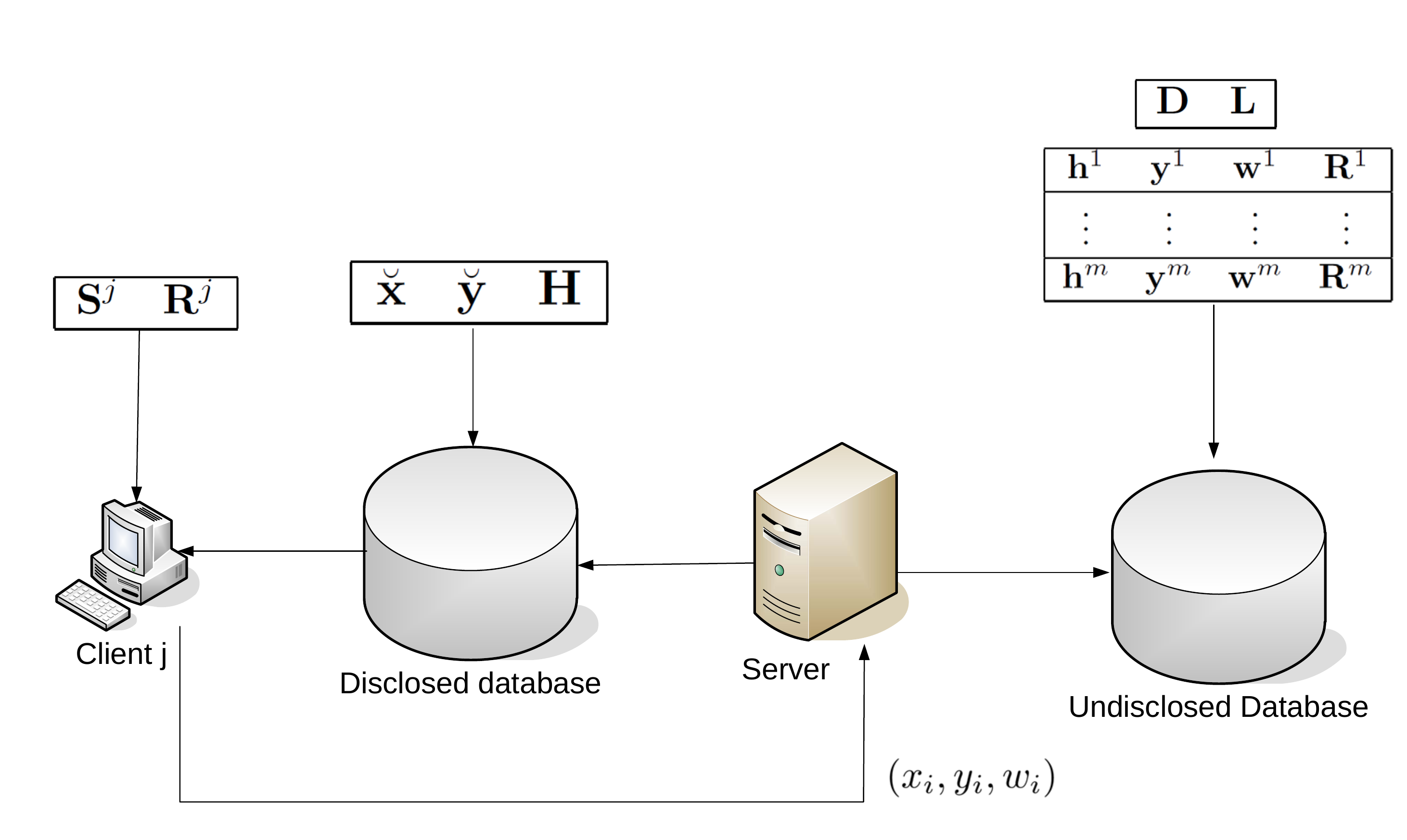}\caption{The client-server scheme.}\label{FIG1}
\end{figure}

Now, we are ready to describe the structure of the client-server algorithm. It is assumed that each client is associated with a different task. The role of the server is twofold:
\begin{enumerate}
    \item \emph{Collecting} triples $(x_i,y_i,w_i)$ (input-output-weight) from the clients and
    updating
        on-line all matrices and coefficients needed to compute estimates for all the tasks.
    \item \emph{Publishing} sufficient information so that any client (task) $j$ can independently compute
        its estimate $\hat{f}_j$, possibly without sending data to the server.
\end{enumerate}

On the other hand, each client $j$ can perform two kind of operations:
\begin{enumerate}
    \item \emph{Sending} triples $(x_i,y_i,w_i)$ to the server.
    \item \emph{Receiving} information from the server sufficient to compute its own estimate $\hat{f}_j$.
\end{enumerate}

\noindent It is required that each client can neither access other clients data nor reconstruct their individual estimates. We have the following scheme:
\begin{itemize}
    \item Undisclosed Information: $\mathbf{h}^j$, $\mathbf{y}^j$, $\mathbf{w}^j$, $\mathbf{R}^j$, for $j\in[m]$.
    \item Disclosed Information: $\breve{\mathbf{x}}$, $\breve{\mathbf{y}}$,  $\mathbf{H}$.
\end{itemize}

\subsection{Server side} \label{sub04-1}

In order to formulate the algorithm in compact form, it is useful to introduce the functions “find”, “ker” and “bias”. Let
\[
A(x) := \left\{i: x_i = x\right\}.
\]

\noindent For any $p,q \in \mathbb{N}$, $x \in X$, $\mathbf{x} \in X^{p}, \mathbf{y} \in X^{q}$, let
\[
\begin{array}{rcl}
\textrm{find}:  X \times X^{p} & \rightarrow & [p+1]\\
\textrm{find}(x,\mathbf{x}) & = & \left\{%
\begin{array}{ll}
p+1, &  A(x)= \O, \\
\min A(x), & A(x) \neq \O. \\
\end{array}%
\right. \\
\textrm{ker}(\cdot,\cdot;K): X^{p}\times X^{q} & \rightarrow &\mathbb{R}^{p\times q}\\
\textrm{ker}\left(\mathbf{x},\mathbf{y};K\right)_{ij} & = & K\left(x_i,y_j\right)\\
\textrm{bias}: X^{p} & \rightarrow & \mathbb{R}^{p\times d}\\
\textrm{bias}\left(\mathbf{x}\right)_{ij} & = & \psi_{j}(x_i).
\end{array}
\]

The complete computational scheme is reported in Algorithm \ref{ALG2}. The initialization is defined by resorting to empty matrices whose manipulation rules can be found in \cite{Nett93}. In particular, $\mathbf{h}^j$, $\mathbf{y}^j$, $\mathbf{w}^j$, $\mathbf{R}^j$, $\mathbf{D}$, $\mathbf{L}$, $\mathbf{M}$, $\breve{\mathbf{x}}$, $\breve{\mathbf{y}}$, $\mathbf{H}$ are all initialized to empty matrix. In this respect, it is assumed that functions “ker” and “bias” return an empty matrix as output, when applied to empty matrices.

\begin{algorithm}
\caption{Server: receive $(x_i, y_i, w_i)$ from client $j$ and update the database.} \label{ALG2}
\begin{algorithmic}[1]
\STATE $s = \textrm{find}\left(x_i,\breve{\mathbf{x}}\right)$
\IF{$(s = n+1)$}
\STATE $n \leftarrow n+1$,
\STATE $\breve{\mathbf{x}} \leftarrow \left(
        \begin{array}{cc}
        \breve{\mathbf{x}}  & x_i \\
        \end{array}
        \right)$,
\STATE $\breve{\mathbf{y}} \leftarrow \left(
        \begin{array}{c}
        \breve{\mathbf{y}} \\
        0 \\
        \end{array}
        \right)$,
\STATE $\mathbf{\overline{k}} \leftarrow \textrm{ker}\left(x_i,\breve{\mathbf{x}};\overline{K}\right)$,
\STATE $\boldsymbol{\psi} \leftarrow \textrm{bias}(x_i)$,
\STATE $\mathbf{r} \leftarrow \textrm{Solution to } \mathbf{L}\mathbf{D}\mathbf{r} = \mathbf{\overline{k}}([n-1])$,
\STATE $\beta \leftarrow \overline{k}_n-\mathbf{r}^T\mathbf{D}\mathbf{r}$,
\STATE $\mathbf{M} \leftarrow \left(
        \begin{array}{c}
        \mathbf{M} \\
        \beta^{-1}\left(\boldsymbol{\psi}-\mathbf{r}^T\mathbf{D}\mathbf{M}\right) \\
        \end{array}
        \right)$,
\STATE $\mathbf{H} \leftarrow \left(%
        \begin{array}{cc}
        \mathbf{H} & \mathbf{0} \\
        \mathbf{0}^T & \beta \\
        \end{array}%
        \right) $
\STATE $\mathbf{D} \leftarrow \left(
        \begin{array}{cc}
        \mathbf{D} & \mathbf{0} \\
        \mathbf{0} & \beta \\
        \end{array}
        \right) $
\STATE $\mathbf{L} \leftarrow \left(%
        \begin{array}{cc}
        \mathbf{L}  & \mathbf{0} \\
        \mathbf{r}^T  & 1 \\
        \end{array}%
        \right)$
\ENDIF
\STATE $p = \textrm{find}\left(x_i,\breve{\mathbf{x}}(\mathbf{h}^j)\right)$
\IF{$(p = \ell_j+1)$}
\STATE $\ell_j \leftarrow \ell_j+1$
\STATE $\mathbf{h}^j \leftarrow \left(
        \begin{array}{cc}
        \mathbf{h}^j & s \\
        \end{array}
        \right)$
\STATE $\mathbf{y}^j \leftarrow \left(
        \begin{array}{c}
        \mathbf{y}^j \\
        y_i \\
        \end{array}
        \right)$
\STATE $\mathbf{w}^j \leftarrow \left(
        \begin{array}{c}
        \mathbf{w}^j \\
        w_i \\
        \end{array}
        \right)$
\STATE $\mathbf{\widetilde{k}} \leftarrow (1-\alpha) \cdot \textrm{ker}\left(x_i,\breve{\mathbf{x}}(\mathbf{h}^j);\widetilde{K}^{j}\right)$,
\STATE $\mathbf{u} \leftarrow \left(
                                \begin{array}{c}
                                  \mathbf{R}^j\mathbf{\widetilde{k}}([\ell_j-1]) \\
                                  -1 \\
                                \end{array}
                              \right)$
\STATE $\gamma \leftarrow 1/\left(\lambda w_i-\mathbf{u}^T \mathbf{\widetilde{k}}\right)$.
\STATE $\mu \leftarrow \gamma \mathbf{u}^T\mathbf{y}^j $,
\STATE $\mathbf{R}^j \leftarrow \left(%
        \begin{array}{cc}
        \mathbf{R}^j   &  \mathbf{0} \\
        \mathbf{0}^{T}  &  0 \\
        \end{array}%
        \right)$
\ELSE
\STATE $\mathbf{u} \leftarrow \mathbf{R}^j(:,p)$,
\STATE $w^j_p \leftarrow w^j_p w_i/\left(w^j_p+w_i\right)$,
\STATE $y^j_p \leftarrow y^j_p+\frac{w^j_p}{w_i}(y_i-y^j_p)$,
\STATE $\gamma \leftarrow \left[\lambda (w^j_p)^2/\left(w_i-w^j_p\right)-R^j_{pp}\right]^{-1}$,
\STATE $\mu \leftarrow w_p^j(y_i-y_{p}^j)/(w_i-w_p^j) +\gamma\mathbf{u}^T\mathbf{y}^j$,
\ENDIF
\STATE $\mathbf{R}^j \leftarrow \mathbf{R}^j +\gamma \mathbf{u}\mathbf{u}^T$
\STATE $\mathbf{v} \leftarrow \mathbf{L}^T(:,\mathbf{h}^j)\mathbf{u}$
\STATE $\mathbf{z} \leftarrow \mathbf{H}\mathbf{v} $,
\STATE $\breve{\mathbf{y}} \leftarrow \breve{\mathbf{y}} + \mu \mathbf{v}$,
\STATE $\mathbf{H} \leftarrow \mathbf{H}-\frac{\mathbf{z}\mathbf{z}^T}{(\alpha \gamma)^{-1}+ \mathbf{v}^T\mathbf{z}}$.
\end{algorithmic}
\end{algorithm}

Algorithm \ref{ALG2} is mainly based on the use of matrix factorizations and matrix manipulation lemmas in the Appendix. The rest of this subsection is an extensive proof devoted to show that Algorithm \ref{ALG2} correctly updates all the relevant quantities when a new triple $(x_i,y_i,w_i)$ becomes available from task $j$. Three cases are possible:
\begin{enumerate}
    \item The input $x_i$ is already among the inputs of task $j$.
    \item The input $x_i$ is not among the inputs of task $j$, but can be found in the common database $\breve{\mathbf{x}}$.
    \item The input $x_i$ is new.
\end{enumerate}

\subsubsection{Case 1: repetition within inputs of task $j$}

The input $x_i$ has been found in $\breve{\mathbf{x}}(\mathbf{h}^j)$, so that it is also present in $\breve{\mathbf{x}}$. Thus, we have
\[
s \neq n+1, \qquad p \neq \ell_j +1,
\]

\noindent and the flow of Algorithm \ref{ALG2} can be equivalently reorganized as in Algorithm \ref{ALG3}.

\begin{algorithm}
\caption{Server (Case 1).} \label{ALG3}
\begin{algorithmic}[1]
\STATE $s = \textrm{find}\left(x_i,\breve{\mathbf{x}}\right)$
\STATE $p = \textrm{find}\left(x_i,\breve{\mathbf{x}}(\mathbf{h}^j)\right)$
\STATE $w^j_p \leftarrow w^j_p w_i/\left(w^j_p+w_i\right)$,
\STATE $y^j_p \leftarrow y^j_p+\frac{w^j_p}{w_i}(y_i-y^j_p)$,
\STATE $\gamma \leftarrow \left[\lambda (w^j_p)^2/\left(w_i-w^j_p\right)-R^j_{pp}\right]^{-1}$,
\STATE $\mathbf{u} \leftarrow \mathbf{R}^j(:,p)$,
\STATE $\mathbf{R}^j \leftarrow \mathbf{R}^j +\gamma \mathbf{u}\mathbf{u}^T$
\STATE $\mu \leftarrow w_p^j(y_i-y_{p}^j)/(w_i-w_p^j) +\gamma\mathbf{u}^T\mathbf{y}^j$,
\STATE $\mathbf{v} \leftarrow \mathbf{L}^T(:,\mathbf{h}^j)\mathbf{u}$
\STATE $\breve{\mathbf{y}} \leftarrow \breve{\mathbf{y}} + \mu \mathbf{v}$,
\STATE $\mathbf{z} \leftarrow \mathbf{H}\mathbf{v} $,
\STATE $\mathbf{H} \leftarrow \mathbf{H}-\frac{\mathbf{z}\mathbf{z}^T}{(\alpha \gamma)^{-1}+ \mathbf{v}^T\mathbf{z}}$.
\end{algorithmic}
\end{algorithm}

Let $r$ denote the number of triples of the type $(x,y_i,w_i)$ belonging to task $j$. These data can be replaced by a single triple $(x,y,w)$ without changing the output of the algorithm. Let
\[
w :=\left(\sum_{i=1}^{r}\frac{1}{w_i}\right)^{-1}, \qquad y := w \sum_{i=1}^{r}\frac{y_i}{w_i},
\]

\noindent The part of the empirical risk regarding these data can be rewritten as
\begin{align*}
&\sum_{i=1}^{r}\frac{\left(y_i-f_j(x)\right)^2}{2w_i}\\
= & \frac{1}{2}\left(\sum_{i=1}^{r}\frac{y_i^2}{w_i}-2f_j(x)\sum_{i=1}^{r}\frac{y_i}{w_i}+f_j(x)^2\sum_{i=1}^{r}\frac{1}{w_i}\right)\\
= & \frac{\left(f_j(x)^2-2f_j(x)y\right)}{2w}+\sum_{i=1}^{r}\frac{y_i^2}{2w_i}\\
= & \frac{\left(y-f_j(x)\right)^2}{2w}+A,
\end{align*}

\noindent where $A$ is a constant independent of $f$.  To recursively update $w$ and $y$ when a repetition is detected, notice that
\begin{align*}
w^{r+1} & =\left(\sum_{i=1}^{r+1}\frac{1}{w_i}\right)^{-1} = \left(\frac{1}{w^{r}}+\frac{1}{w_{r+1} }\right)^{-1}=\frac{w^r w_{r+1}}{w^r+w_{r+1}},\\
y^{r+1} & = w^{r+1}\sum_{i=1}^{r+1}\frac{y_i}{w_i} =\left(\frac{1}{w^{r}}+\frac{1}{w_{r+1}}\right)^{-1} \left(\frac{y^{r}}{w^{r}}+\frac{y_{r+1}}{w_{r+1}}\right)\\
& = y^r+\frac{w^r}{w_i}(y_i-y^r).
\end{align*}

\noindent By applying these formulas to the $p$-th data of task $j$, lines $3,4$ of Algorithm \ref{ALG3} are obtained. To check that $\mathbf{R}^j$ is correctly updated by lines $5,6,7$ of Algorithm \ref{ALG3} just observe that, taking into account the definition of $\mathbf{R}^j$ and applying Lemma \ref{LEM1}, we have:
\begin{align*}
& \left( (1-\alpha) \mathbf{\widetilde{K}}^j(\mathbf{h}^j,\mathbf{h}^j) +\lambda \mathbf{W}(\mathbf{h}^j,\mathbf{h}^j) - \frac{\lambda \mathbf{e}_p\mathbf{e}_p^T (w_{p}^j)^2}{\left(w_i-w_{p}^j\right)} \right)^{-1}\\
& = \mathbf{R}^j  +\frac{\mathbf{R}^j  \mathbf{e}_p \mathbf{e}_p^T \mathbf{R}^j }{\lambda (w_{p}^j)^2/\left(w_i-w_{p}^j\right)-\mathbf{e}_p^T\mathbf{R}^j \mathbf{e}_p} \\
& = \mathbf{R}^j+\gamma \mathbf{u}\mathbf{u}^T.
\end{align*}

\noindent Consider now the update of $\breve{\mathbf{y}}$.  Since $\mathbf{y}^j$ has already been updated, the previous $\mathbf{y}^j$ is given by
\[
\mathbf{y}^j-\mathbf{e}_p\Delta y_{p}^j,
\]

\noindent where the variation $\Delta y_{p}^j$ of $y_{p}^j$ can be expressed as
\[
\Delta y_{p}^j = w_p^j\frac{y_i-y_{p}^j}{w_i-w_p^j}.
\]

\noindent Recalling the definition of $\breve{\mathbf{y}}$ in Algorithm \ref{ALG1}, and line 7 of Algorithm \ref{ALG3}, we have
\[
\breve{\mathbf{y}} \leftarrow \mathbf{L}^T\left(\sum_{k\neq j}^{m}\mathbf{I}(:,\mathbf{h}^k)\mathbf{R}^k\mathbf{y}^k+ \mathbf{I}(:,\mathbf{h}^j)\left(\mathbf{R}^j+\gamma\mathbf{u}\mathbf{u}^T\right)\mathbf{y}^j\right).
\]

\noindent By adding and subtracting $\Delta y_{p}^j\mathbf{e}_p$, using the definition of $\mu$ in line 8,
\begin{align*}
& \left(\mathbf{R}^j+\gamma\mathbf{u}\mathbf{u}^T\right)\mathbf{y}^j \\
& = \mathbf{R}^j\left(\mathbf{y}^j-\Delta y_{p}^j\mathbf{e}_p\right)+ \left(\Delta y_{p}^j +\gamma \left(\mathbf{u}^T\mathbf{y}^j\right)\right)\mathbf{u}\\
& = \mathbf{R}^j\left(\mathbf{y}^j-\Delta y_{p}^j\mathbf{e}_p\right) + \mu \mathbf{u}.
\end{align*}

\noindent Hence,
\[
\breve{\mathbf{y}} \leftarrow \breve{\mathbf{y}} + \mu \mathbf{L}^T(:,\mathbf{h}^j)  \mathbf{u}
\]
\noindent By defining $\mathbf{v}$ as in line 9 of Algorithm \ref{ALG3}, the update of line 10 is obtained. Finally, we show that $\mathbf{H}$ is correctly updated. Let
\[
\mathbf{F} := \alpha \mathbf{L}^T\mathbf{P}^T\mathbf{R}\mathbf{P}\mathbf{L}.
\]

\noindent Then, from the definition of $\mathbf{H}$ it follows that
\[
\mathbf{H} = \left(\mathbf{D}^{-1}+\mathbf{F} \right)^{-1}.
\]

\noindent In view of lines 7,9 of Algorithm \ref{ALG3},
\[
\mathbf{F} \leftarrow \mathbf{F} + \alpha \gamma \mathbf{v} \mathbf{v}^T,
\]
\noindent so that
\[
\mathbf{H} \leftarrow \left(\mathbf{H}^{-1}  + \alpha \gamma \mathbf{v} \mathbf{v}^T\right)^{-1}.
\]

\noindent By Lemma \ref{LEM1}, lines 11, 12 are obtained.

\subsubsection{Case 2: repetition in $\breve{\mathbf{x}}$.}

Since $x_i$ belongs to $\breve{\mathbf{x}}$ but not to $\breve{\mathbf{x}}(\mathbf{h}^j)$, we have
\[
s \neq n+1, \qquad p = \ell_j +1.
\]

\noindent The flow of Algorithm \ref{ALG2} can be organized as in Algorithm \ref{ALG4}.

\begin{algorithm}
\caption{Server (Case 2)} \label{ALG4}
\begin{algorithmic}[1]
\STATE $s = \textrm{find}\left(x_i,\breve{\mathbf{x}}\right)$
\STATE $p = \textrm{find}\left(x_i,\breve{\mathbf{x}}(\mathbf{h}^j)\right)$
\STATE $\ell_j \leftarrow \ell_j+1$
\STATE $\mathbf{h}^j \leftarrow \left(
        \begin{array}{cc}
        \mathbf{h}^j & s \\
        \end{array}
        \right)$
\STATE $\mathbf{y}^j \leftarrow \left(
        \begin{array}{c}
        \mathbf{y}^j \\
        y_i \\
        \end{array}
        \right)$
\STATE $\mathbf{w}^j \leftarrow \left(
        \begin{array}{c}
        \mathbf{w}^j \\
        w_i \\
        \end{array}
        \right)$
\STATE $\mathbf{\widetilde{k}} \leftarrow (1-\alpha) \cdot \textrm{ker}\left(x_i,\breve{\mathbf{x}}(\mathbf{h}^j);\widetilde{K}^{j}\right)$,
\STATE $\mathbf{u} \leftarrow \left(
                                \begin{array}{c}
                                  \mathbf{R}^j\mathbf{\widetilde{k}}([\ell_j-1]) \\
                                  -1 \\
                                \end{array}
                              \right)$
\STATE $\gamma \leftarrow 1/\left(\lambda w_i-\mathbf{u}^T \mathbf{\widetilde{k}}\right)$.
\STATE $\mathbf{R}^j \leftarrow \left(%
        \begin{array}{cc}
        \mathbf{R}^j   &  \mathbf{0} \\
        \mathbf{0}^{T}  &  0 \\
        \end{array}%
        \right) +\gamma \mathbf{u}\mathbf{u}^T$
\STATE $\mathbf{v} \leftarrow \mathbf{L}^T(:,\mathbf{h}^j)\mathbf{u}$
\STATE $\mu \leftarrow \gamma \mathbf{u}^T\mathbf{y}^j $,
\STATE $\breve{\mathbf{y}} \leftarrow \breve{\mathbf{y}} + \mu \mathbf{v}$,
\STATE $\mathbf{z} \leftarrow \mathbf{H}\mathbf{v} $,
\STATE $\mathbf{H} \leftarrow \mathbf{H}-\frac{\mathbf{z}\mathbf{z}^T}{(\alpha \gamma)^{-1}+ \mathbf{v}^T\mathbf{z}}$.
\end{algorithmic}
\end{algorithm}

\noindent First, vectors $\mathbf{h}^j$, $\mathbf{y}^{j}$ and $\mathbf{w}^j$ must be properly enlarged as in lines 3-6. Recalling the definition of $\mathbf{R}^j$, we have:
\[
(\mathbf{R}^j)^{-1} \leftarrow
\left(
\begin{array}{cc}
(\mathbf{R}^j)^{-1} & \mathbf{\widetilde{k}}([\ell_j-1])\\
\mathbf{\widetilde{k}}([\ell_j-1])^T   & \widetilde{k}_{\ell_j}+\lambda w_i \\
\end{array}
\right)
\]

\noindent The update for $\mathbf{R}^j$ in lines 7-10 is obtained by applying Lemma \ref{LEM2} with $\mathbf{A} = (\mathbf{R}^j)^{-1}$.

Consider now the update of $\breve{\mathbf{y}}$. Recall that $\mathbf{h}^j$ and $\mathbf{y}^j$ have already been updated. By the definition of $\breve{\mathbf{y}}$ and in view of line 10 of Algorithm \ref{ALG4}, we have
\begin{align*}
\breve{\mathbf{y}} & \leftarrow \mathbf{L}^T\sum_{k\neq j}^{m}\mathbf{I}(:,\mathbf{h}^k)\mathbf{R}^k\mathbf{y}^k\\
& + \mathbf{L}^T\mathbf{I}(:,\mathbf{h}^j)\left[\left(%
        \begin{array}{cc}
        \mathbf{R}^j   &  \mathbf{0} \\
        \mathbf{0}^{T}  &  0 \\
        \end{array}%
        \right)+\gamma\mathbf{u}\mathbf{u}^T\right]\mathbf{y}^j\\
& = \breve{\mathbf{y}}+ \gamma (\mathbf{u}^T\mathbf{y}^j) \mathbf{L}^T\mathbf{I}(:,\mathbf{h}^j) \mathbf{u}.
\end{align*}

\noindent The update in lines 11-13 immediately follows. Finally, the update in lines 14-15 for $\mathbf{H}$ is obtained by applying Lemma \ref{LEM2} as in Case 1.

\subsubsection{Case 3: $x_i$ is a new input.}

\begin{algorithm}
\caption{Server (Case 3)} \label{ALG5}
\begin{algorithmic}[1]
\STATE $n \leftarrow n+1$
\STATE $\breve{\mathbf{x}} \leftarrow \left(
        \begin{array}{cc}
        \breve{\mathbf{x}}  & x_i \\
        \end{array}
        \right)$.
\STATE $\mathbf{\overline{k}} \leftarrow \textrm{ker}\left(x_i,\breve{\mathbf{x}};\overline{K}\right)$,
\STATE $\mathbf{r} \leftarrow \textrm{Solution to } \mathbf{L}\mathbf{D}\mathbf{r} = \mathbf{\overline{k}}([n-1])$,
\STATE $\beta \leftarrow \overline{k}_n-\mathbf{r}^T\mathbf{D}\mathbf{r}$,
\STATE $\boldsymbol{\psi} \leftarrow \textrm{bias}(x_i)$,
\STATE $\mathbf{M} \leftarrow \left(
        \begin{array}{c}
        \mathbf{M} \\
        \beta^{-1}\left(\boldsymbol{\psi}-\mathbf{r}^T\mathbf{D}\mathbf{M}\right) \\
        \end{array}
        \right)$,
\STATE $\mathbf{D} \leftarrow \left(
        \begin{array}{cc}
        \mathbf{D} & \mathbf{0} \\
        \mathbf{0} & \beta \\
        \end{array}
        \right)$
\STATE $\mathbf{L} \leftarrow \left(%
        \begin{array}{cc}
        \mathbf{L}  & \mathbf{0} \\
        \mathbf{r}^T  & 1 \\
        \end{array}%
        \right)$
\STATE $\breve{\mathbf{y}} \leftarrow \left(
        \begin{array}{c}
        \breve{\mathbf{y}} \\
        0 \\
        \end{array}
        \right)$
\STATE $\mathbf{H} \leftarrow \left(%
        \begin{array}{cc}
        \mathbf{H} & \mathbf{0} \\
        \mathbf{0}^T & \beta \\
        \end{array}%
        \right)$
\STATE $\textrm{Call Algorithm}$ \ref{ALG4}.
\end{algorithmic}
\end{algorithm}

Since $x_i$ is a new input, we have
\[
s = n+1, \qquad p = \ell_j +1.
\]

\noindent The flow of Algorithm \ref{ALG2} can be reorganized as in Algorithm \ref{ALG5}. The final part of Algorithm \ref{ALG5} coincides with Algorithm \ref{ALG4}. However, the case of new input also requires updating factors $\mathbf{D}$ and $\mathbf{L}$ and matrix $\mathbf{M}$. Assume that $\overline{K}$ is strictly positive so that $\mathbf{D}$ is diagonal and $\mathbf{L}$ is lower triangular. If $\overline{K}$ is not strictly positive, other kinds of decompositions can be used. In particular, for the linear kernel $\overline{K}(x_1,x_2) = \left<x_1, x_2\right>$ over $\mathbb{R}^r$, $\mathbf{D}$ and $\mathbf{L}$ can be taken, respectively, equal to the identity and $\breve{\mathbf{x}}$. Recalling that $\breve{\mathbf{K}} = \mathbf{L}\mathbf{D}\mathbf{L}^T$, we have
\begin{align*}
\breve{\mathbf{K}}
& \leftarrow
\left(
\begin{array}{cc}
\breve{\mathbf{K}} & \mathbf{\overline{k}}([n-1])\\
\mathbf{\overline{k}}([n-1])^T   & \overline{k}_n\\
\end{array}
\right) \\
& = \left(
\begin{array}{cc}
\mathbf{L}\mathbf{D}\mathbf{L}^T & \mathbf{\overline{k}}([n-1])\\
\mathbf{\overline{k}}([n-1])^T   & \overline{k}_n\\
\end{array}
\right)\\
 & = \left(%
        \begin{array}{cc}
        \mathbf{L}  & \mathbf{0} \\
        \mathbf{r}^T  & 1 \\
        \end{array}%
        \right)
        \left(
        \begin{array}{cc}
        \mathbf{D} & \mathbf{0} \\
        \mathbf{0} & \beta \\
        \end{array}
        \right)
        \left(%
        \begin{array}{cc}
        \mathbf{L}  & \mathbf{0} \\
        \mathbf{r}^T  & 1 \\
        \end{array}%
        \right)^T,
\end{align*}

\noindent with $\mathbf{r}$ and $\mathbf{\beta}$ as in lines 4-5.

Concerning $\mathbf{M}$, recall from Assumption \ref{ASS1} that
\[
\mathbf{L}\mathbf{D}\mathbf{M} = \breve{\boldsymbol{\Psi}}.
\]

\noindent The relation must remain true by substituting the updated quantities. Indeed, after the update in lines 6-9, we have
\begin{align*}
\mathbf{L}\mathbf{D}\mathbf{M} & \leftarrow \left(%
        \begin{array}{cc}
        \mathbf{L}\mathbf{D}  & \mathbf{0} \\
        \mathbf{r}^T\mathbf{D}  & \beta \\
        \end{array}%
        \right) \left(
        \begin{array}{c}
        \mathbf{M} \\
        \beta^{-1}\left(\boldsymbol{\psi}-\mathbf{r}^T\mathbf{D}\mathbf{M}\right) \\
        \end{array}
        \right)\\
        & =
        \left(
        \begin{array}{c}
        \mathbf{L}\mathbf{D}\mathbf{M} \\
        \mathbf{r}^T\mathbf{D}\mathbf{M}+ \boldsymbol{\psi}-\mathbf{r}^T\mathbf{D}\mathbf{M}\\
        \end{array}
        \right)\\
        & = \left(
        \begin{array}{c}
        \breve{\boldsymbol{\Psi}} \\
        \boldsymbol{\psi} \\
        \end{array}
        \right).
\end{align*}

\noindent Finally, it is easy to see that updates for $\breve{\mathbf{y}}$ and $\mathbf{H}$ are similar to that of previous Case 2, once the enlargements in lines 10-11 are made.

\subsection{Client side}  \label{sub04-2}

To obtain coefficients $\mathbf{a}$ by Algorithm \ref{ALG1}, access to undisclosed data $\mathbf{h}^j$, $\mathbf{y}^j$, $\mathbf{R}^j$ is required. Nevertheless, as shown next, each client can compute its own estimate $\hat{f}_j$ without having access to the undisclosed data. It is not even necessary to know the overall number $m$ of tasks, nor their “individual kernels” $\widetilde{K}^j$: all the required information is contained in the disclosed quantities $\breve{\mathbf{x}}$, $\breve{\mathbf{y}}$ and $\mathbf{H}$. From the client point of view, knowledge of $\breve{\mathbf{x}}$ is equivalent to the knowledge of $\breve{\mathbf{K}}$ and $\breve{\boldsymbol{\Psi}}$. In turn, also $\mathbf{L}$, $\mathbf{D}$ and $\mathbf{M}$ can be computed using the factorization (\ref{E06}) and the definition of $\mathbf{M}$ in Assumption \ref{ASS1}. As mentioned in the introduction, two kind of clients are considered.
\begin{itemize}
  \item An \emph{active} client $j$ sends its own data to the server. This kind of client can request both the disclosed information and its individual coefficients $\mathbf{a}^j$ (Algorithm \ref{ALG6}).
  \item A \emph{passive} client $j$ does not send its data. In this case, the server is not able to compute $\mathbf{a}^j$. This kind of client can only request the disclosed information, and must run a local version of the server to obtain $\mathbf{a}^j$ (Algorithm \ref{ALG7}).
\end{itemize}

\noindent The following Theorem ensures that vector $\breve{\mathbf{a}}$ can be computed by knowing only disclosed data.

\begin{thm}\label{THM2}
Given $\breve{\mathbf{x}}$, $\breve{\mathbf{y}}$ and $\mathbf{H}$, the condensed coefficients vector $\breve{\mathbf{a}}$ can be computed by solving the linear system
\[
\mathbf{D}\mathbf{L}^T \breve{\mathbf{a}} = \mathbf{H}\left(\breve{\mathbf{y}}+ \mathbf{M}\mathbf{b}\right)-\mathbf{D}\mathbf{M}\mathbf{b}.
\]
\end{thm}

Once the disclosed data and vector $\breve{\mathbf{a}}$ have been obtained, each client still needs the individual coefficients vector $\mathbf{a}^j$ in order to perform predictions for its own task. While an active client can simply receive such vector from the server, a passive client must compute it independently. Interestingly, it turns out that $\mathbf{a}^j$ can be computed by knowing only disclosed data together with private data of task $j$. Indeed, line 11 of Algorithm \ref{ALG1} decouples with respect to the different tasks:
\[
\mathbf{a}^j \leftarrow \mathbf{R}^j\left[\mathbf{y}^j-\alpha\mathbf{L}(\mathbf{h}^j,:)\left(\mathbf{z}+\mathbf{H}\mathbf{M}\mathbf{b}\right)\right].
\]
\noindent This is the key feature that allows a passive client to perform predictions without disclosing its private data and exploiting the information contained in all the other datasets.

\begin{algorithm}
\caption{(Active client $j$) Receive $\breve{\mathbf{x}}$, $\breve{\mathbf{y}}$, $\mathbf{H}$ and $\mathbf{a}^j$ and evaluate $\breve{\mathbf{a}}$, $\mathbf{b}$} \label{ALG6}
\begin{algorithmic}[1]
\FOR{$i = 1:n$}
\STATE $\mathbf{\overline{k}} \leftarrow \textrm{ker}\left(\breve{x}_i,\breve{\mathbf{x}}([i]);\overline{K}\right)$,
\STATE $\mathbf{r} \leftarrow \textrm{Solution to } \mathbf{L}\mathbf{D}\mathbf{r} = \mathbf{\overline{k}}([i-1])$,
\STATE $\beta \leftarrow \overline{k}_i-\mathbf{r}^T\mathbf{D}\mathbf{r}$,
\STATE $\mathbf{D} \leftarrow \left(
        \begin{array}{cc}
        \mathbf{D} & \mathbf{0} \\
        \mathbf{0} & \beta \\
        \end{array}
        \right)$,
\STATE $\mathbf{L} \leftarrow \left(%
        \begin{array}{cc}
        \mathbf{L}  & \mathbf{0} \\
        \mathbf{r}^T  & 1 \\
        \end{array}%
        \right)$,
\STATE $\boldsymbol{\psi} \leftarrow \textrm{bias}(x_i)$,
\STATE $\mathbf{M} \leftarrow \left(
        \begin{array}{c}
        \mathbf{M} \\
        \beta^{-1}\left(\boldsymbol{\psi}-\mathbf{r}^T\mathbf{D}\mathbf{M}\right) \\
        \end{array}
        \right)$,
\ENDFOR
\STATE $\mathbf{z} \leftarrow \mathbf{H}\breve{\mathbf{y}}$
\STATE $\mathbf{b} \leftarrow \textrm{Solution to} \left(\mathbf{M}^T(\mathbf{D}- \mathbf{H})\mathbf{M}\right)\mathbf{b} = \mathbf{M}^T\mathbf{z}$,
\STATE $\breve{\mathbf{a}} \leftarrow \textrm{Solution to} \left(\mathbf{D}\mathbf{L}^T \right)\breve{\mathbf{a}} = \mathbf{z}+(\mathbf{H}-\mathbf{D})\mathbf{M}\mathbf{b}.$
\end{algorithmic}
\end{algorithm}

\begin{algorithm}
\caption{(Passive client $j$) Receive $\breve{\mathbf{x}}$, $\breve{\mathbf{y}}$ and $\mathbf{H}$ and evaluate $\breve{\mathbf{a}}$, $\mathbf{b}$ and $\mathbf{a}^j$} \label{ALG7}
\begin{algorithmic}[1]
\FOR{$i = 1:n$}
\STATE $\mathbf{\overline{k}} \leftarrow \textrm{ker}\left(\breve{x}_i,\breve{\mathbf{x}}([i]);\overline{K}\right)$,
\STATE $\mathbf{r} \leftarrow \textrm{Solution to } \mathbf{L}\mathbf{D}\mathbf{r} = \mathbf{\overline{k}}([i-1])$,
\STATE $\beta \leftarrow \overline{k}_i-\mathbf{r}^T\mathbf{D}\mathbf{r}$,
\STATE $\mathbf{D} \leftarrow \left(
        \begin{array}{cc}
        \mathbf{D} & \mathbf{0} \\
        \mathbf{0} & \beta \\
        \end{array}
        \right)$,
\STATE $\mathbf{L} \leftarrow \left(%
        \begin{array}{cc}
        \mathbf{L}  & \mathbf{0} \\
        \mathbf{r}^T  & 1 \\
        \end{array}%
        \right)$,
\STATE $\boldsymbol{\psi} \leftarrow \textrm{bias}(x_i)$,
\STATE $\mathbf{M} \leftarrow \left(
        \begin{array}{c}
        \mathbf{M} \\
        \beta^{-1}\left(\boldsymbol{\psi}-\mathbf{r}^T\mathbf{D}\mathbf{M}\right) \\
        \end{array}
        \right)$,
\ENDFOR
\FOR{$i =1: \ell_j$}
\STATE Run a local version of Algorithm \ref{ALG2} with $(x_{ij},y_{ij},w_{ij})$.
\ENDFOR
\STATE $\mathbf{z} \leftarrow \mathbf{H}\breve{\mathbf{y}}$
\STATE $\mathbf{b} \leftarrow \textrm{Solution to} \left(\mathbf{M}^T(\mathbf{D}- \mathbf{H})\mathbf{M}\right)\mathbf{b} = \mathbf{M}^T\mathbf{z}$,
\STATE $\breve{\mathbf{a}} \leftarrow \textrm{Solution to} \left(\mathbf{D}\mathbf{L}^T \right)\breve{\mathbf{a}} = \mathbf{z}+(\mathbf{H}-\mathbf{D})\mathbf{M}\mathbf{b}.$
\STATE $\mathbf{a}^j \leftarrow \mathbf{R}^j\left[\mathbf{y}^j-\alpha\mathbf{L}(\mathbf{h}^j,:)\left(\mathbf{z}+\mathbf{H}\mathbf{M}\mathbf{b}\right)\right]$.
\end{algorithmic}
\end{algorithm}

\section{Illustrative example: music recommendation} \label{sec05}

In this section, the proposed algorithm is applied to a simulated music recommendation problem, in order to predict preferences of several virtual users with respect to a set of artists. Artist data were obtained from the May 2005 AudioScrobbler Database dump  \footnote{ \url{http://www-etud.iro.umontreal.ca/~bergstrj/audioscrobbler_data.html}} which is the last dump released by AudioScrobbler/LastFM under Creative Commons license. LastFM is an internet radio that provides individualized broadcasts based on user preferences. The database dump includes users playcounts and artists names so that it is possible to rank artists according to global number of playcounts. After sorting artists according to decreasing playcounts, 489 top ranking artists were selected. The input space $X$ is therefore a set of $489$ artists, i.e.
\[
X = \left\{\textrm{Bob Marley}, \textrm{Madonna}, \textrm{Michael Jackson}, ... \right\}.
\]
\noindent The tasks are associated with user preference functions. More precisely, normalized preferences of user $j$ over the entire set of artists are expressed by functions $s_j: X \rightarrow [0, 1]$ defined as
\[
s_j(x_i) = \frac{1}{1+e^{-f_j(x_i)/2}}.
\]
\noindent where $f_j:X \rightarrow \mathbb{R}$ are the tasks to be learnt.

\begin{figure}
\centering
\includegraphics[width=0.8\linewidth]{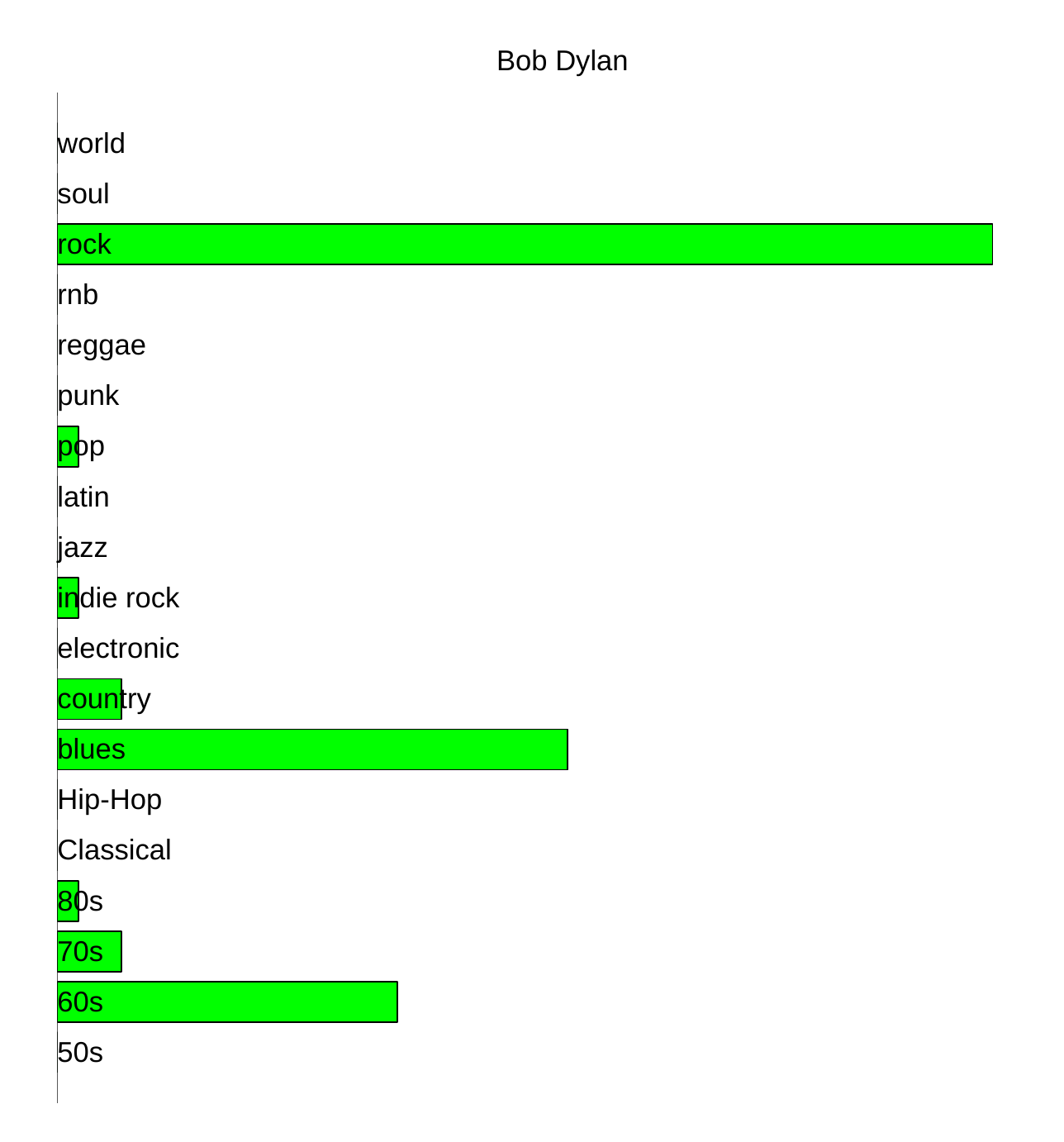}\caption{Example of artist tagging}\label{FIG2}
\end{figure}

The simulated music recommendation system relies on music type classification expressed by means of tags (rock, pop, ...). In particular, the 19 main tags of LastFM \footnote{http://www.lastfm.com} were considered. The $i$-th artist is associated with a vector $\mathbf{z}_i \in [0, 1]^{19}$ of 19 \emph{tags}, whose values were obtained by querying the LastFM site on September 22, 2008. In Figure \ref{FIG2}, the list of the tags considered in this experiment, together with an example of artist's tagging are reported. Vectors $\mathbf{z}_i$ have been normalized to lie on the unit hyper-sphere, i.e. $\|\mathbf{z}_i\|_{2} = 1$. The input space data (artists together with their normalized tags) are available for download \footnote{\url{http://www-dimat.unipv.it/~dinuzzo/files/mrdata.zip}}.

Tag information was used to build a mixed-effect kernel over $X$. More precisely, $\overline{K}$ is a Gaussian RBF kernel and $\widetilde{K}^k = \widetilde{K}$ are linear kernels:
\begin{align*}
\overline{K}(x_i(\mathbf{z}_i),x_j(\mathbf{z}_j)) & =  e^{1-\|\mathbf{z}_i-\mathbf{z}_j\| /2} = e^{\mathbf{z}_i^T\mathbf{z}_j},\\
\widetilde{K}(x_i(\mathbf{z}_i),x_j(\mathbf{z}_j)) & = \mathbf{z}_i^T\mathbf{z}_j.
\end{align*}

The above kernels were employed to generate synthetic users. First, an “average user” was generated by drawing a function $\overline{f}:X \rightarrow \mathbb{R}$ from a Gaussian process with zero mean and auto-covariance $\overline{K}$. Then, $m=3000$ virtual user's preferences were generated as
\[
f_j = 0.25\overline{f} + 0.75\widetilde{f}_j,
\]
\noindent where $\widetilde{f}_j$ are drawn from a Gaussian process with zero mean and auto-covariance $\widetilde{K}$. For the $j$-th virtual user, $\ell_j = 5$ artists $x_{ij}$ were uniformly randomly sampled from the input space $X$, and corresponding noisy outputs $y_{ij}$ generated as
\[
y_{ij} = f_j(x_{ij}) + \epsilon_{ij},
\]
\noindent where $\epsilon_{ij}$ are i.i.d. Gaussian errors with zero mean and standard deviation $\sigma = 0.01$.  The learned preference function $\hat{s}_j$ is
\[
\hat{s}_j(x_i) = \frac{1}{1+e^{-\hat{f}_j(x_i)/2}},
\]

\noindent where $\hat{f}_j$ is estimated using the algorithm described in the paper. Performances are evaluated by both the average root mean squared error
\[
\textrm{RMSE} = \sqrt{\frac{1}{m|X|}\sum_{i=1}^{|X|}\sum_{j=1}^{m}(s_j(x_i)-\hat{s}_j(x_i))^2},
\]

\noindent and the average number of hits over the top 20 ranked artists, defined as
\begin{align*}
\textrm{TOP20HITS} &= \frac{1}{m}\sum_{i=1}^{m}\textrm{hits20}_j,\\
\textrm{hits20}_j &:=\left| \textrm{top20}(s_j) \cap \textrm{top20}(\hat{s}_j)\right|,
\end{align*}

\noindent where $\textrm{top20}:\mathcal{H} \rightarrow X^{20}$ returns the sorted vector of 20 inputs with highest scores, measured by a function $s:X \rightarrow [0, 1], s \in \mathcal{H}$.

\begin{figure}
\centering
\includegraphics[width=1\linewidth]{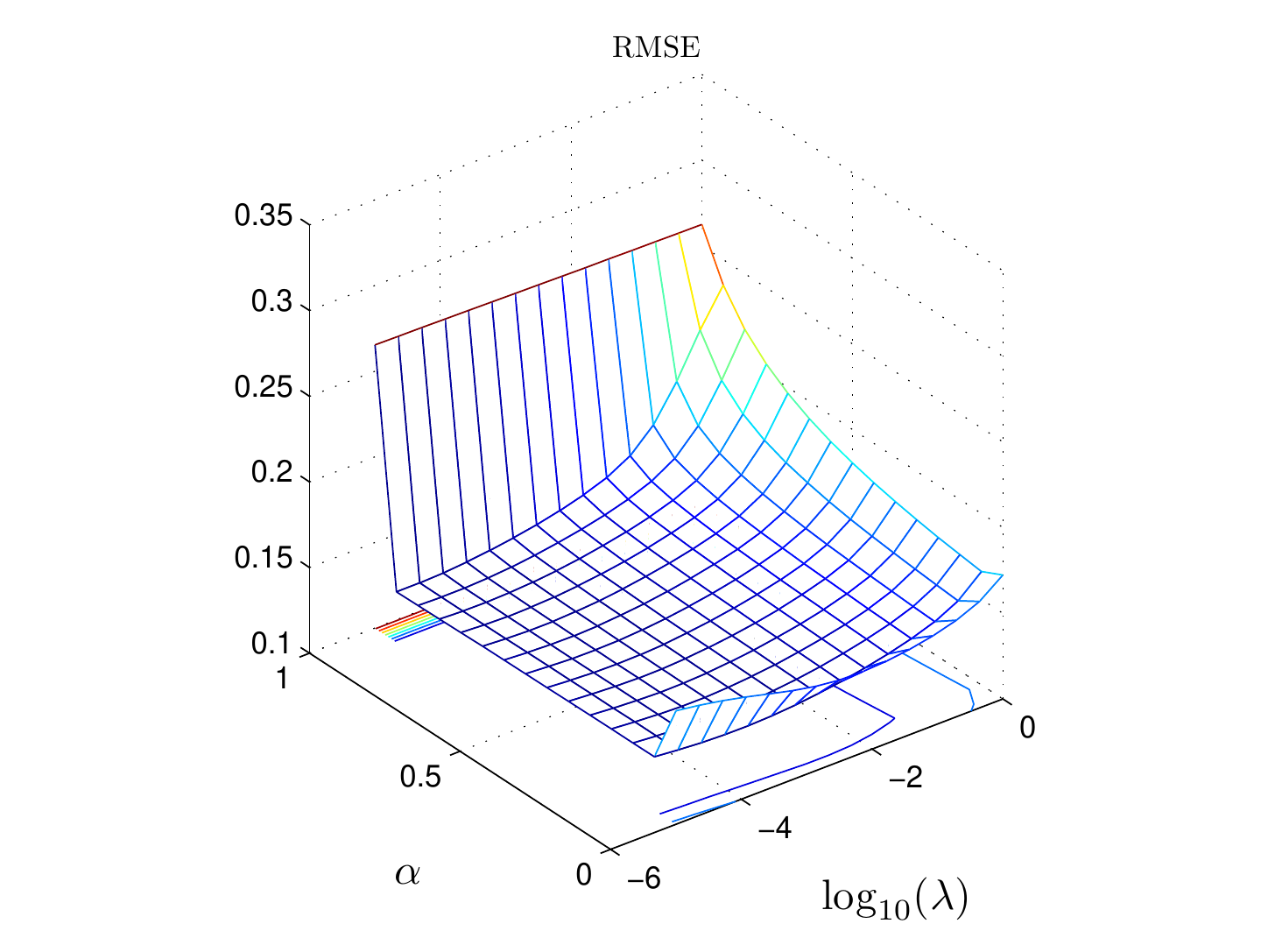}
\includegraphics[width=1\linewidth]{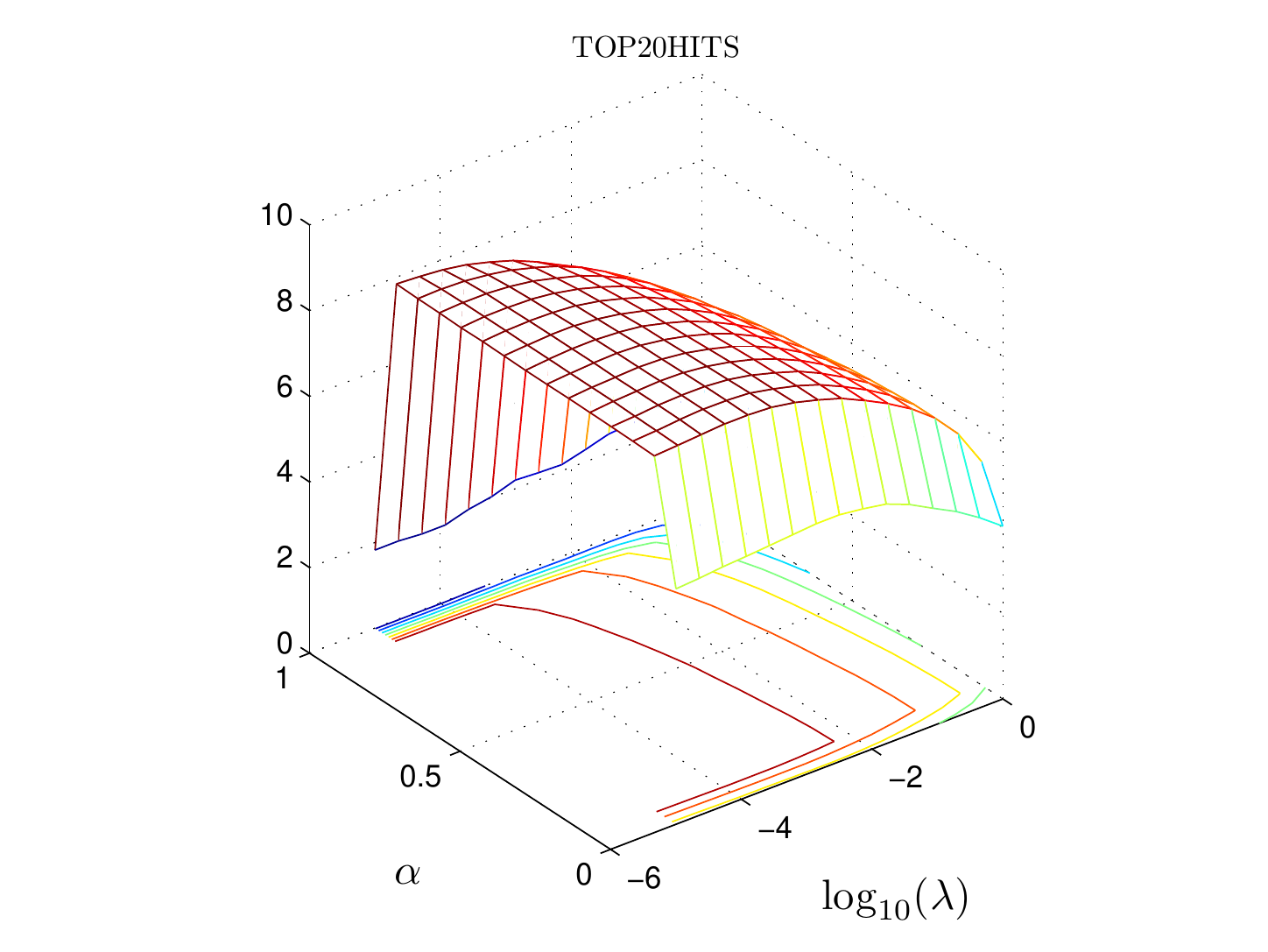}

\caption{Average $\textrm{TOP20HITS}$ and $\textrm{RMSE}$ against $\alpha$ and $\lambda$.}\label{FIG3}
\end{figure}

\begin{figure}
\centering
\includegraphics[width=0.9\linewidth]{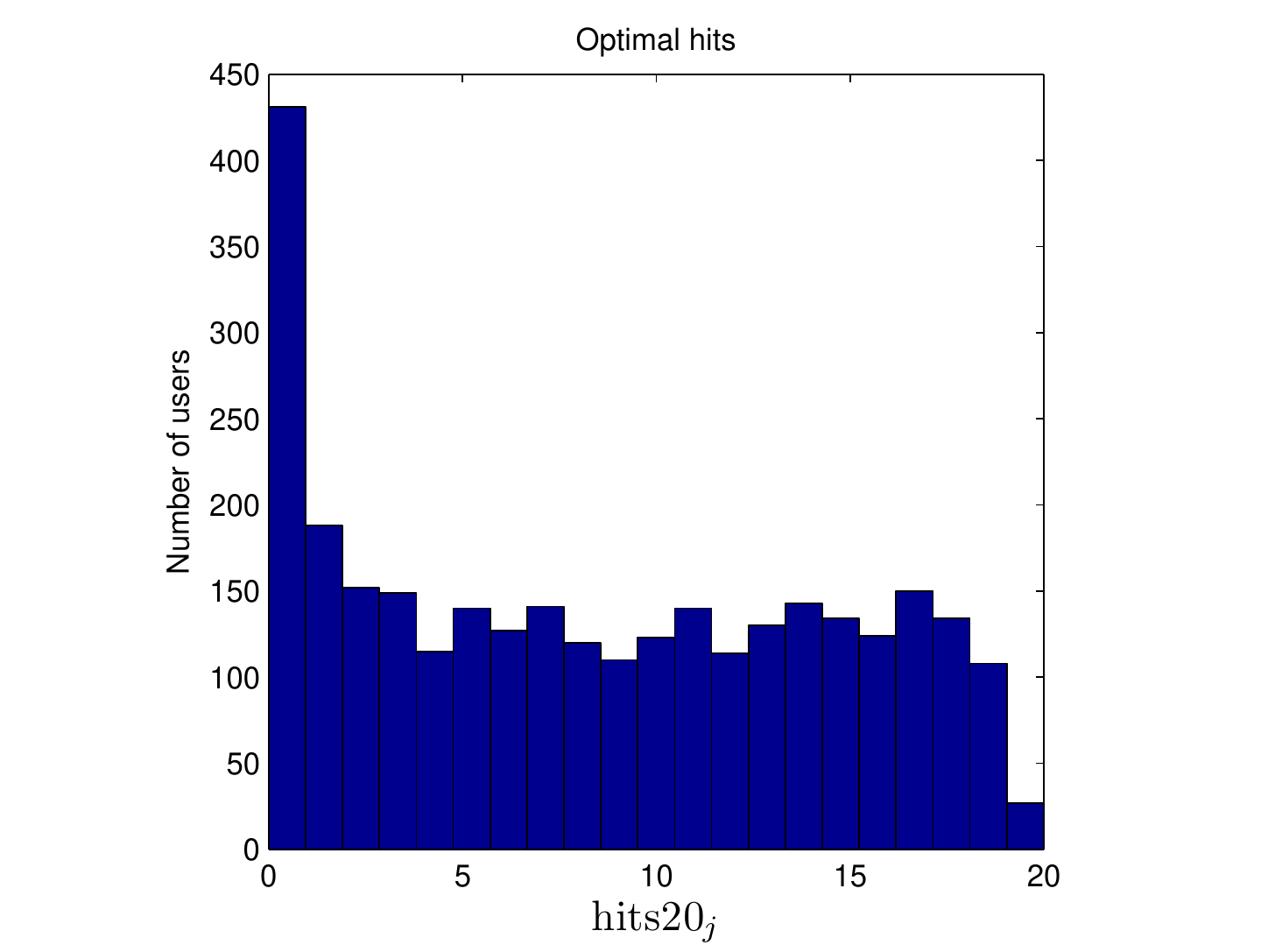}\caption{Distribution of $\textrm{hits20}_j$ in correspondence with $\alpha^*$ and $\lambda^*$ achieving optimal $RMSE$.}\label{FIG4}
\end{figure}

\begin{figure}
\centering
\includegraphics[width=1\linewidth]{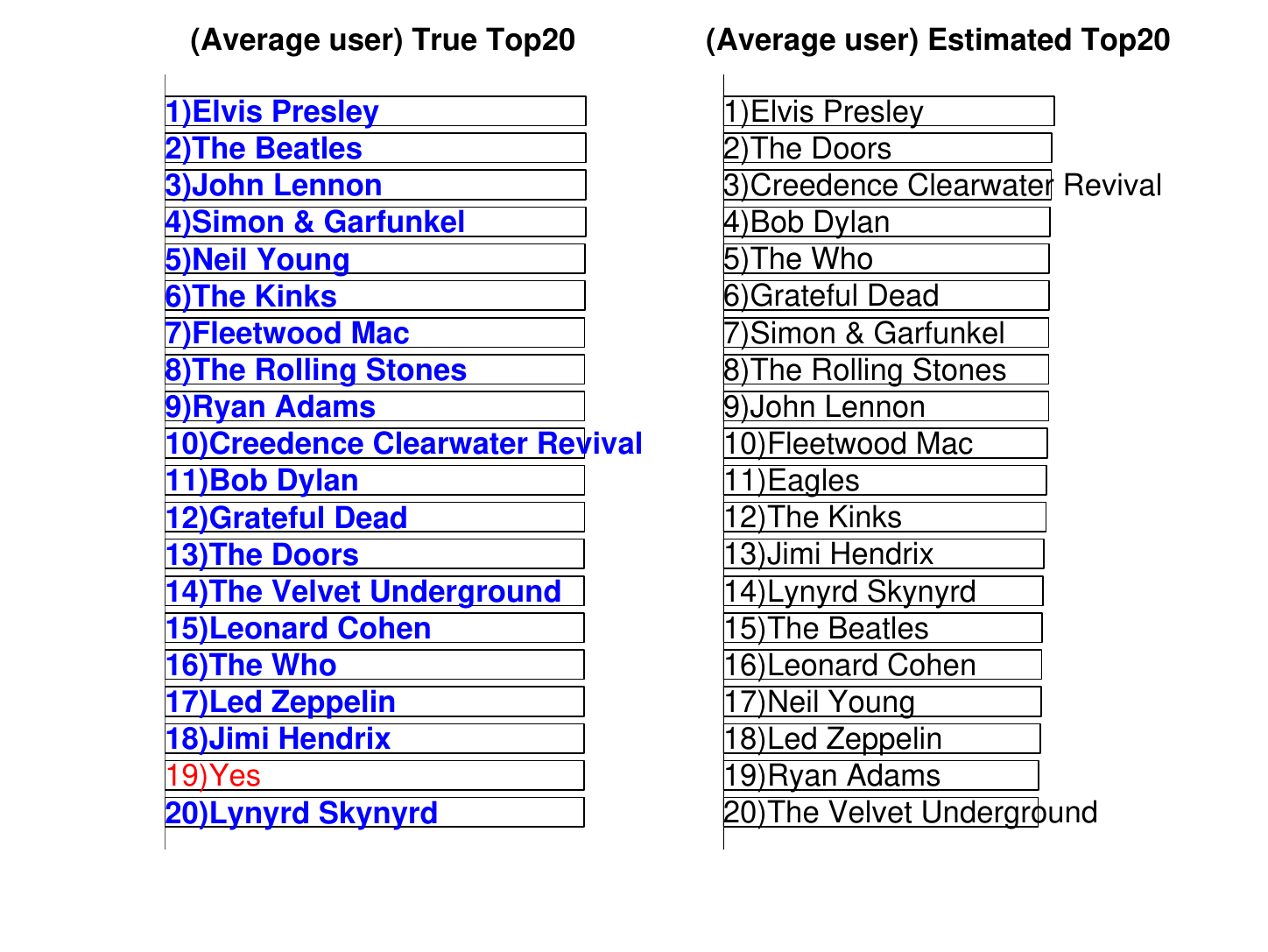}\caption{True and estimated Top20 for the “average user”.}\label{FIG5}
\end{figure}

\begin{figure}
\centering
\includegraphics[width=1\linewidth]{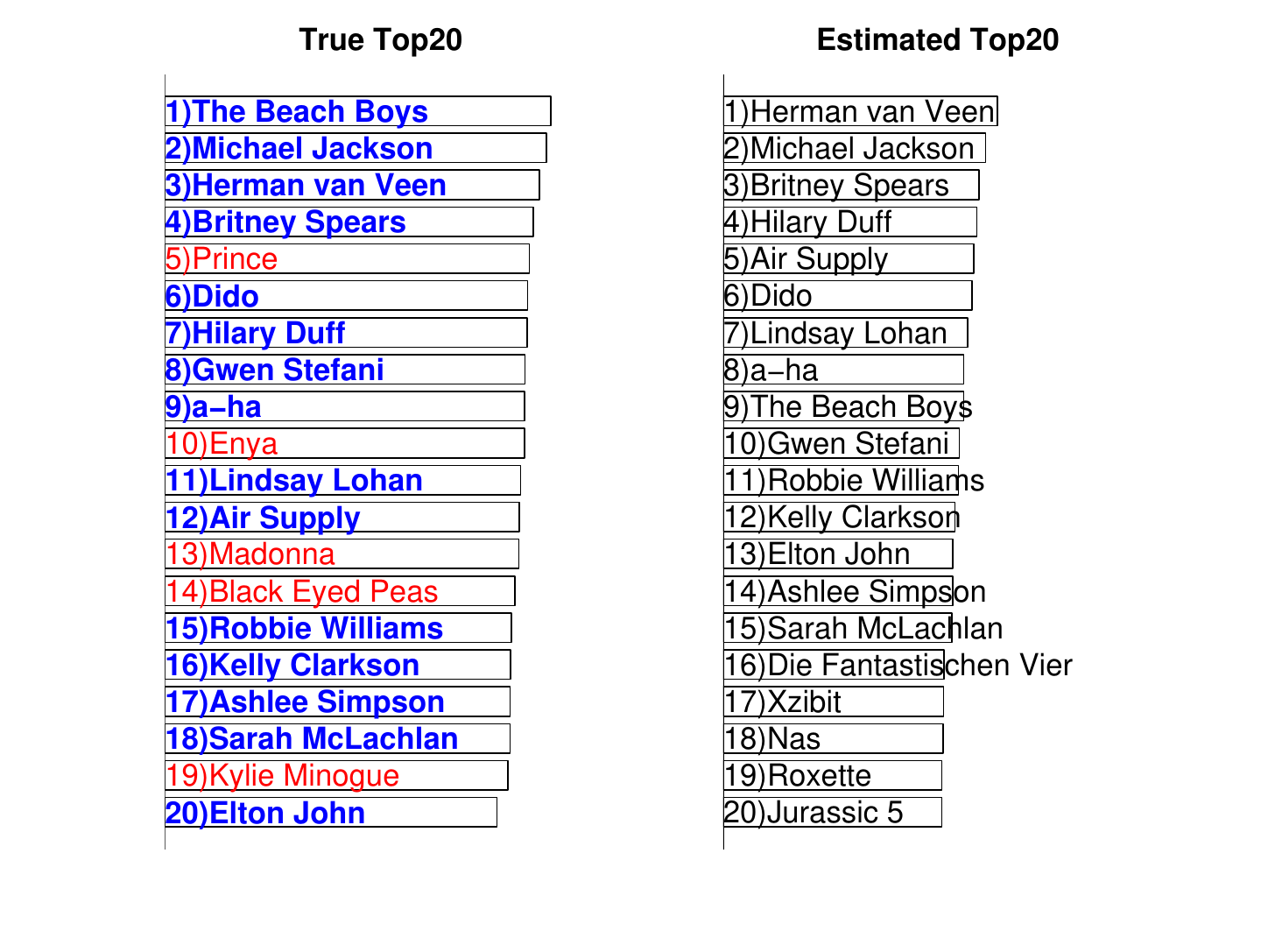}
\includegraphics[width=1\linewidth]{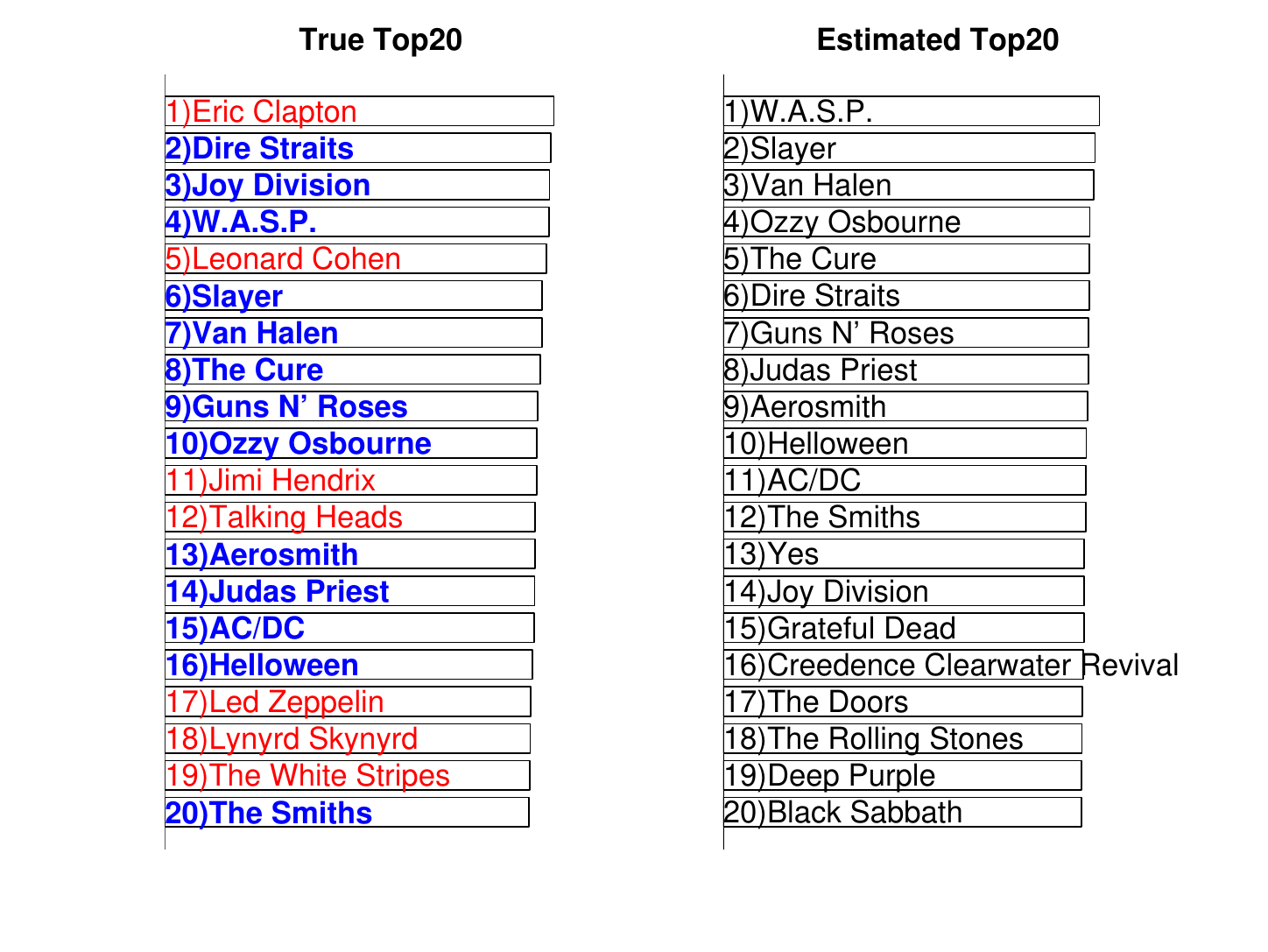}
\caption{True and estimated Top20 for two representative users.}\label{FIG6}
\end{figure}

Learning was performed for 15 values of the shrinking parameter $\alpha$ linearly spaced in $\left[0, 1\right]$ and 15 values of the regularization parameter $\lambda$ logarithmically spaced in the interval $\left[10^{-7}, 10^{0}\right]$, see Figure \ref{FIG3}. The multi-task approach, i.e. $0 < \alpha < 1$ outperforms both the separate ($\alpha = 0$) and pooled ($\alpha = 1$) approaches. Interestingly, performances remain fairly stable for a range of values of $\alpha$. Figure \ref{FIG4} shows the distribution of $\textrm{hits20}_j$ over the 3000 users in correspondence with values of $\alpha^*$ and $\lambda^*$ achieving the optimal $\textrm{RMSE}$. Although $\alpha^* =  0.0714$ and $\lambda^* = 3.1623  \cdot 10^{-4}$ were selected so as to minimize the $\textrm{RMSE}$, remarkably good performances are obtained also with respect to the \textrm{TOP20HITS} score which is 8.3583 (meaning that on the average 8.3583 artists among the top-20 are correctly retrieved). Finally, true and estimated top-20 hits are reported for the average user (Figure \ref{FIG5}) and two representative users (Figure \ref{FIG6}). Artists of the true top-20 that are correctly retrieved in the estimated top-20 are reported in bold-face.

Concerning the computational burden, it is worth observing that without exploiting the presence of repeated inputs and the mixed-effect structure of the kernel, the complexity of a naive approach would be of the order of the cube of the overall number of examples, that is $(5 \cdot 3000)^3$. Conversely, the complexity of the approach proposed in the paper scales with $n^3 m$, where $n$ is the number of unique inputs and $m$ the number of tasks (in our example, $n$ is bounded by the cardinality of the input set $|X| = 489$, and $m =3000$).

\section{Conclusions} \label{sec06}

Recent studies have highlighted the potentialities of kernel methods applied to multi-task learning, but their effective implementation involve the solution of  architectural and complexity issues. In this paper, emphasis is posed on the architecture with reference to learning from distributed datasets. For a general class of kernels with a “mixed-effect” structure it is shown that the optimal solution can be given a collaborative client-server architecture that enjoys favorable computational and confidentiality properties. By interacting with the server, each client can solve its own estimation task while taking advantage of all the data from the other clients without having any direct access to them. Client's privacy is preserved, since both active and passive clients are allowed by the architecture. The former are those that agree to send their data to the server while the latter only exploit information from the server without disclosing their private data. The proposed architecture has several potential applications ranging from biomedical data analysis (where privacy issues are crucial) to web data mining. An illustrative example is given by the simulated music recommendation system discussed in the paper.

\section*{Acknowledgments}
This work has been partially supported by MIUR Project \emph{Artificial pancreas: physiological models, control algorithms and clinical test} and PRIN Project \emph{Metodi e algoritmi innovativi per la stima Bayesiana e l'identificazione e il controllo adattativo e distribuito}. The authors would like to thank Pietro De Nicolao for preprocessing data used in the numerical example.

\appendix \label{app01}

\noindent Recall the following two lemmas on matrix inversions, see e.g. \cite{Golub96}.

\begin{lem}[Sherman-Morrison-Woodbury]\label{LEM1}

Let $\mathbf{A} \in \mathbb{R}^{n \times n}$, $\mathbf{B} \in \mathbb{R}^{m \times m}$ be two nonsingular
matrix, $\mathbf{U} \in \mathbb{R}^{n \times m}$, $\mathbf{V} \in \mathbb{R}^{m \times n}$ such that $(\mathbf{A}
+ \mathbf{U}\mathbf{B}\mathbf{V})$ is nonsingular. Then, matrix
\[
\mathbf{E} := \left(\mathbf{B}^{-1}+ \mathbf{V}\mathbf{A}^{-1}\mathbf{U} \right)
\]
\noindent is nonsingular, and
\[
\left(\mathbf{A} + \mathbf{U}\mathbf{B}\mathbf{V}\right)^{-1} = \mathbf{A}^{-1}- \mathbf{A}^{-1} \mathbf{U} \mathbf{E}^{-1}\mathbf{V}\mathbf{A}^{-1}.
\]
\end{lem}

\medskip

\begin{lem}[Schur]\label{LEM2}
Suppose that matrix
\[
\mathbf{A} = \left(%
\begin{array}{cc}
  \mathbf{B} & \mathbf{C} \\
  \mathbf{C}^T & \mathbf{D} \\
\end{array}%
\right)  \in \mathbb{R}^{(n+m) \times (n+m)}
\]

\noindent is nonsingular, with $\mathbf{B} \in \mathbb{R}^{n \times n}$. Then,
\[
\mathbf{E}^{-1} := \left(\mathbf{C}^T \mathbf{B}^{-1} \mathbf{C} - \mathbf{D}\right)
\]
\noindent is nonsingular and
\[
\mathbf{A}^{-1} = \left(%
\begin{array}{cc}
  \mathbf{B}^{-1}-\mathbf{B}^{-1} \mathbf{C} \mathbf{E}\mathbf{C}^T \mathbf{B}^{-1} & \mathbf{B}^{-1} \mathbf{C} \mathbf{E} \\
  \mathbf{E}^T \mathbf{C}^T \mathbf{B}^{-1} & -\mathbf{E}^T  \\
\end{array}%
\right).
\]
\end{lem}

\medskip

\noindent \textbf{Proof of Theorem \ref{THM1}} Let $\mathbf{R}^j$, $\mathbf{R}$ be defined as in line 1-5 of Algorithm \ref{ALG1}, and observe that
\[
\mathbf{R}^{-1}  = (1-\alpha)\sum_{j=1}^{m}\mathbf{I}(:,\mathbf{k}^j)\mathbf{\widetilde{K}}^j(\mathbf{k}^j,\mathbf{k}^j)\mathbf{I}(\mathbf{k}^j,:) + \lambda \mathbf{W}
\]

\noindent  Consider the back-fitting formulation (\ref{E04}), (\ref{E05}) of the linear system (\ref{E03}). By Lemma \ref{LEM1}, we have:
\begin{align*}
\left(\mathbf{K} + \lambda \mathbf{W} \right)^{-1} & = \left(\alpha \mathbf{\overline{K}} + \mathbf{R}^{-1}\right)^{-1}\\
& = \left(\alpha \mathbf{P}\mathbf{L}\mathbf{D}\mathbf{L}^T\mathbf{P}^T + \mathbf{R}^{-1}\right)^{-1}\\
& = \mathbf{R} - \alpha \mathbf{R}\mathbf{P}\mathbf{L}\mathbf{H}\mathbf{L}^T\mathbf{P}^T\mathbf{R}.
\end{align*}

\noindent Let $\mathbf{F} := \alpha \mathbf{L}^T\mathbf{P}^T\mathbf{R}\mathbf{P}\mathbf{L}$, and observe that
\[
\mathbf{D}\mathbf{F}\mathbf{H}= \mathbf{H}\mathbf{F}\mathbf{D}= \mathbf{D}- \mathbf{H}.
\]

\noindent In the following, we exploit the following relationship:
\begin{align*}
& \alpha \mathbf{L}^T\mathbf{P}^T\left(\mathbf{K} + \lambda \mathbf{W} \right)^{-1}\mathbf{P}\mathbf{L} \\
= & \alpha\mathbf{L}^T\mathbf{P}^T\left(\mathbf{R}-\alpha \mathbf{R}\mathbf{P}\mathbf{L}\mathbf{H}\mathbf{L}^T\mathbf{P}^T\mathbf{R}\right)\mathbf{L}\mathbf{P}\\
= & \mathbf{F}-\mathbf{F}\mathbf{H}\mathbf{F}.
\end{align*}

\noindent Consider the case $\alpha \neq 0$. Then, in view of the previous relationship, recalling that $\boldsymbol{\Psi} = \mathbf{P}\breve{\boldsymbol{\Psi}} = \mathbf{P}\mathbf{L}\mathbf{D}\mathbf{M}$, we have
\begin{align*}
& \alpha \boldsymbol{\Psi}^T\left(\mathbf{K} + \lambda \mathbf{W} \right)^{-1} \boldsymbol{\Psi}\\
= & \alpha \mathbf{M}^T \mathbf{D}^T\mathbf{L}^T\mathbf{P}^T \left(\mathbf{K} + \lambda \mathbf{W} \right)^{-1} \mathbf{P}\mathbf{L}\mathbf{D}\mathbf{M}\\
= & \mathbf{M}^T\mathbf{D}\left(\mathbf{F}-\mathbf{F}\mathbf{H}\mathbf{F}\right)\mathbf{D}\mathbf{M}\\
= &\mathbf{M}^T\left(\mathbf{D}-\mathbf{D}\mathbf{F}\mathbf{H}\right)\mathbf{F}\mathbf{D}\mathbf{M}\\
= & \mathbf{M}^T \mathbf{H}\mathbf{F}\mathbf{D}\mathbf{M}\\
= & \mathbf{M}^T (\mathbf{D}- \mathbf{H})\mathbf{M},
\end{align*}

\noindent and
\begin{align*}
& \boldsymbol{\Psi}^T\left(\mathbf{K}+\lambda \mathbf{W}\right)^{-1}\mathbf{y}\\
= & \mathbf{M}^T\mathbf{D} \mathbf{L}^T\mathbf{P}^T\left(\mathbf{R} -\alpha \mathbf{R}\mathbf{P}\mathbf{L}\mathbf{H}\mathbf{L}^T\mathbf{P}^T\mathbf{R}\right)\mathbf{y}\\
= & \mathbf{M}^T\left(\mathbf{D}-\mathbf{D}\mathbf{F}\mathbf{H}\right)\mathbf{L}^T\mathbf{P}^T\mathbf{R}\mathbf{y}\\
= & \mathbf{M}^T\mathbf{H}\breve{\mathbf{y}}.
\end{align*}

\noindent Then, line 10 of Algorithm \ref{ALG1} follows from
(\ref{E04}). Observe that
\[
\left(\mathbf{K} + \lambda \mathbf{W} \right)\mathbf{a} = \alpha \mathbf{\overline{K}}\mathbf{a}+\mathbf{R}^{-1}\mathbf{a}.
\]

\noindent Then, from (\ref{E05}) we have
\[
\mathbf{a}=\mathbf{R}\left[\mathbf{y}-\alpha\mathbf{P}\mathbf{L}\left(\mathbf{D}\mathbf{L}^T\mathbf{P}^T\mathbf{a}+\mathbf{D}\mathbf{M}\mathbf{b}\right)\right].
\]

\noindent Now,
\begin{align*}
& \mathbf{D}\mathbf{L}^T\mathbf{P}^T\mathbf{a} + \mathbf{D}\mathbf{M}\mathbf{b} \\
= & \mathbf{D}\mathbf{L}^T\mathbf{P}^T\left(\mathbf{K} + \lambda \mathbf{W} \right)^{-1}\left(\mathbf{y}-\alpha \mathbf{P}\mathbf{L}\mathbf{D}\mathbf{M}\mathbf{b}\right)+\mathbf{D}\mathbf{M}\mathbf{b}\\
= & \left(\mathbf{D}-\mathbf{D}\mathbf{F}\mathbf{H}\right)\left(\mathbf{L}^T\mathbf{P}^T\mathbf{R}\mathbf{y}-\mathbf{F}\mathbf{D}\mathbf{M}\mathbf{b}\right)+\mathbf{D}\mathbf{M}\mathbf{b}\\
= & \mathbf{H}\mathbf{L}^T\mathbf{P}^T\mathbf{R}\mathbf{y}-\mathbf{H}\mathbf{F}\mathbf{D}\mathbf{M}\mathbf{b}+\mathbf{D}\mathbf{M}\mathbf{b}\\
= & \mathbf{H}\left(\breve{\mathbf{y}}+ \mathbf{M}\mathbf{b}\right).
\end{align*}

\noindent Hence, we obtain line 11 of Algorithm \ref{ALG1}. Finally, for $\alpha = 0$, we have $\mathbf{H} = \mathbf{D}$ so that the thesis follows.

\medskip

\noindent \textbf{Proof of Theorem \ref{THM2}}
Let $\mathbf{F}:= \lambda \mathbf{L}^T\mathbf{P}^T\mathbf{T}\mathbf{P}\mathbf{L}$. Recalling the expression of $\mathbf{a}$, $\breve{\mathbf{y}}$, $\mathbf{H}$ in Algorithm \ref{ALG1}, we have
\begin{align*}
\mathbf{D}\mathbf{L}^T \breve{\mathbf{a}} & = \mathbf{D}\mathbf{L}^T \mathbf{P}^T \mathbf{a}\\
& = \mathbf{D}\mathbf{L}^T \mathbf{P}^T \mathbf{R}\left[\mathbf{y}-\lambda \mathbf{P}\mathbf{L}\mathbf{H}\left(\breve{\mathbf{y}}+ \mathbf{M}\mathbf{b}\right)\right]\\
& = \mathbf{D}\left[\breve{\mathbf{y}}-\mathbf{F}\mathbf{H}\left(\breve{\mathbf{y}}+ \mathbf{M}\mathbf{b}\right)\right]\\
& = \left(\mathbf{D}-\mathbf{D}\mathbf{F}\mathbf{H}\right)\breve{\mathbf{y}}- \mathbf{D}\mathbf{F}\mathbf{H}\mathbf{M}\mathbf{b}\\
& = \mathbf{H}\breve{\mathbf{y}}-\left(\mathbf{D}-\mathbf{H}\right)\mathbf{M}\mathbf{b} \\
& = \mathbf{H}\left(\breve{\mathbf{y}}+ \mathbf{M}\mathbf{b}\right)-\mathbf{D}\mathbf{M}\mathbf{b}.
\end{align*}

\end{document}